\documentclass{bmvc2k}

\usepackage{graphicx}
\usepackage{amsmath}
\usepackage{amssymb}
\usepackage{algpseudocode}
\usepackage{algorithm}
\usepackage{algorithmicx}
\usepackage{epstopdf}
\usepackage{comment}
\usepackage{tabu}
\usepackage{bm}
\usepackage{multicol}
\usepackage[table]{xcolor}
\usepackage{tabularx}
\usepackage[bookmarks=false]{hyperref}

\DeclareMathOperator*{\argmin}{\arg\!\min}
\DeclareMathOperator*{\argmax}{\arg\!\max}

\title{Searching for Objects using Structure in Indoor Scenes}

\addauthor{Varun K. Nagaraja}{varun@umiacs.umd.edu}{1}
\addauthor{Vlad I. Morariu}{morariu@umiacs.umd.edu}{1}
\addauthor{Larry S. Davis}{lsd@umiacs.umd.edu}{1}

\addinstitution{
 University of Maryland\\
College Park, MD. USA.
}

\runninghead{Nagaraja \bmvaEtAl}{Searching for Objects using Structure in Indoor Scenes}

\def\etal{\emph{et al}\bmvaOneDot}

\begin{document}

\maketitle

\begin{abstract}
To identify the location of objects of a particular class, a passive computer vision system generally processes all the regions in an image to finally output few regions. However, we can use structure in the scene to search for objects without processing the entire image. We propose a search technique that sequentially processes image regions such that the regions that are more likely to correspond to the query class object are explored earlier. We frame the problem as a Markov decision process and use an imitation learning algorithm to learn a search strategy. Since structure in the scene is essential for search, we work with indoor scene images as they contain both unary scene context information and object-object context in the scene. We perform experiments on the NYU-depth v2 dataset and show that the unary scene context features alone can achieve a significantly high average precision while processing only 20-25\% of the regions for classes like \textit{bed} and \textit{sofa}. By considering object-object context along with the scene context features, the performance is further improved for classes like \textit{counter}, \textit{lamp}, \textit{pillow} and \textit{sofa}.
\end{abstract}

\section{Introduction}
The current prevalent object detection framework \cite{Girshick2014} is a pipeline of two main stages: the object proposal stage and the feature extraction/classification stage. Object proposals are image regions that with high probability significantly overlap with an object, irrespective of object class. Features are extracted from object proposals and then a label is predicted. Even with high quality object proposals, the typical number of proposals considered by the feature extraction stage ranges from hundreds to tens of thousands for high resolution imagery. 

Consider the situation where a computer vision system needs to identify the presence or location of a particular object in an image. In a passive computer vision system, if we ask a specific question like ``Where is the table in this room?", it would process all the region proposals in the image to detect a table instance. Such a vision system does not exploit the structure in the scene to efficiently process the image. 
Our goal is to locate objects of interest in an image by processing as few image regions as possible using scene structure. We build on a region proposal module that generates candidate regions and a region classification module that predicts the class label for a region. The generic strategy is to sequentially process image regions such that the regions that are more likely to correspond to the object of interest are explored earlier. At each step, we use the labels of the explored regions and spatial context to predict the likelihood that each unexplored region is an instance of the target class. We then select a few regions with highest likelihood, obtain the class label from the region classification module and add them to the explored set. The process is repeated with the updated set of explored regions. 

We frame our sequential exploration problem as a Markov Decision Process (MDP) and use a reinforcement learning technique to learn an optimal search policy.
However, it is challenging to manually specify a reward function for the search policy. The true reward function is unknown for our sequential exploration problem since the underlying distribution from which a spatial arrangement of objects in an image is generated is unknown, analogous to a game generated by a hidden emulator \cite{Mnih2015}. But we have access to an oracle's actions in the individual images. 
Learning an optimal policy in such situations is known as imitation learning \cite{Abbeel2004} where an oracle predicts the actions it would take at a state and the search policy learns to imitate the oracle and predict similar actions. The oracle in our image exploration problem selects the next set of regions to explore based on the groundtruth labels. We use the DAgger algorithm of Ross \etal \cite{Ross2011} that trains a classifier as the search policy on a dataset of features extracted at states and actions taken by the oracle (labels), where the states are generated by running the policy iteratively over the training data.

\begin{figure*}
\centering
\footnotesize
\subfigure[Ranked sequence obtained from an object proposal technique.]{
{\includegraphics[width=0.24\linewidth, trim=1cm 1cm 0cm 0.5cm,clip=true]{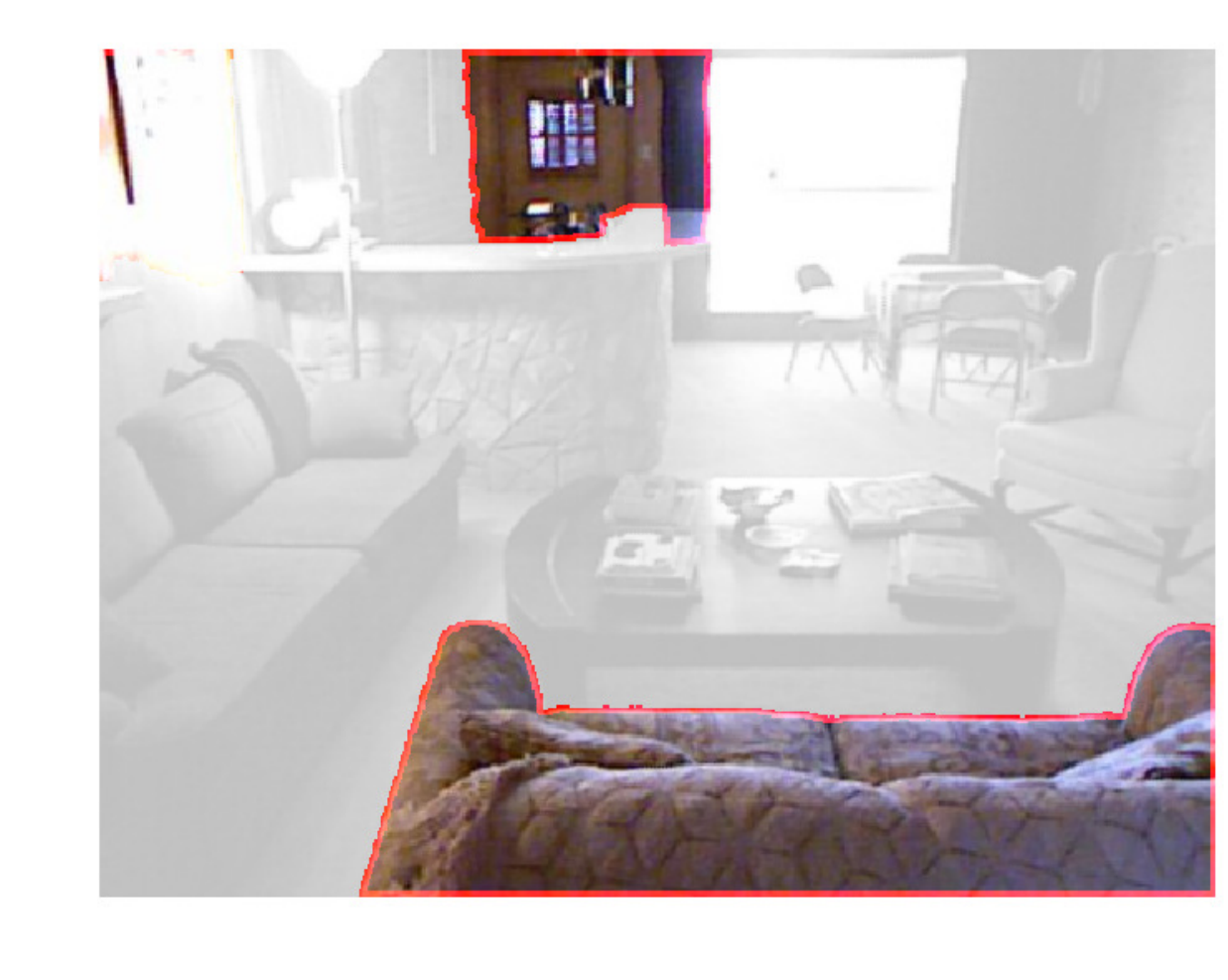}}
\includegraphics[width=0.24\linewidth, trim=1cm 1cm 0cm 0.5cm,clip=true]{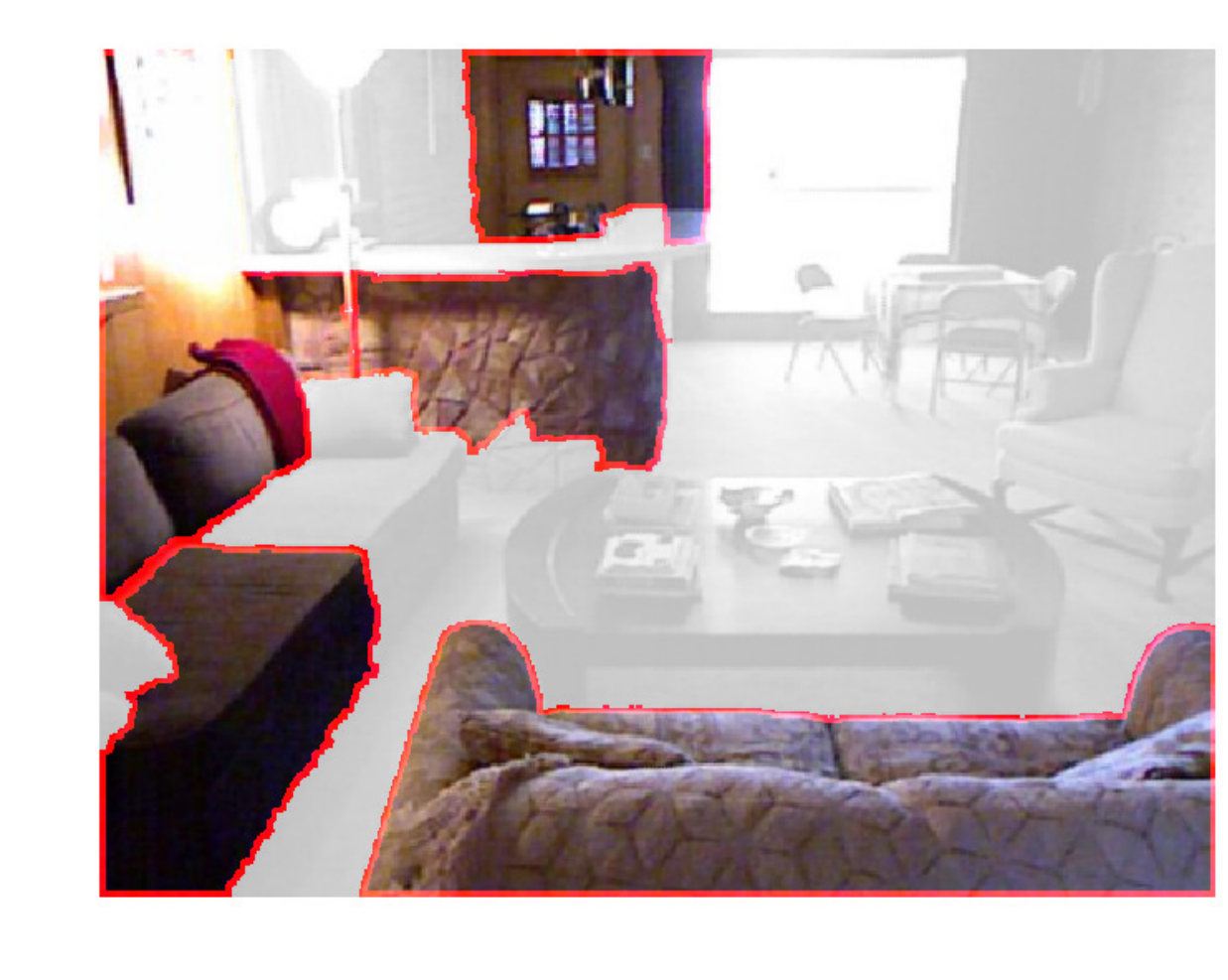}
\includegraphics[width=0.24\linewidth, trim=1cm 1cm 0cm 0.5cm,clip=true]{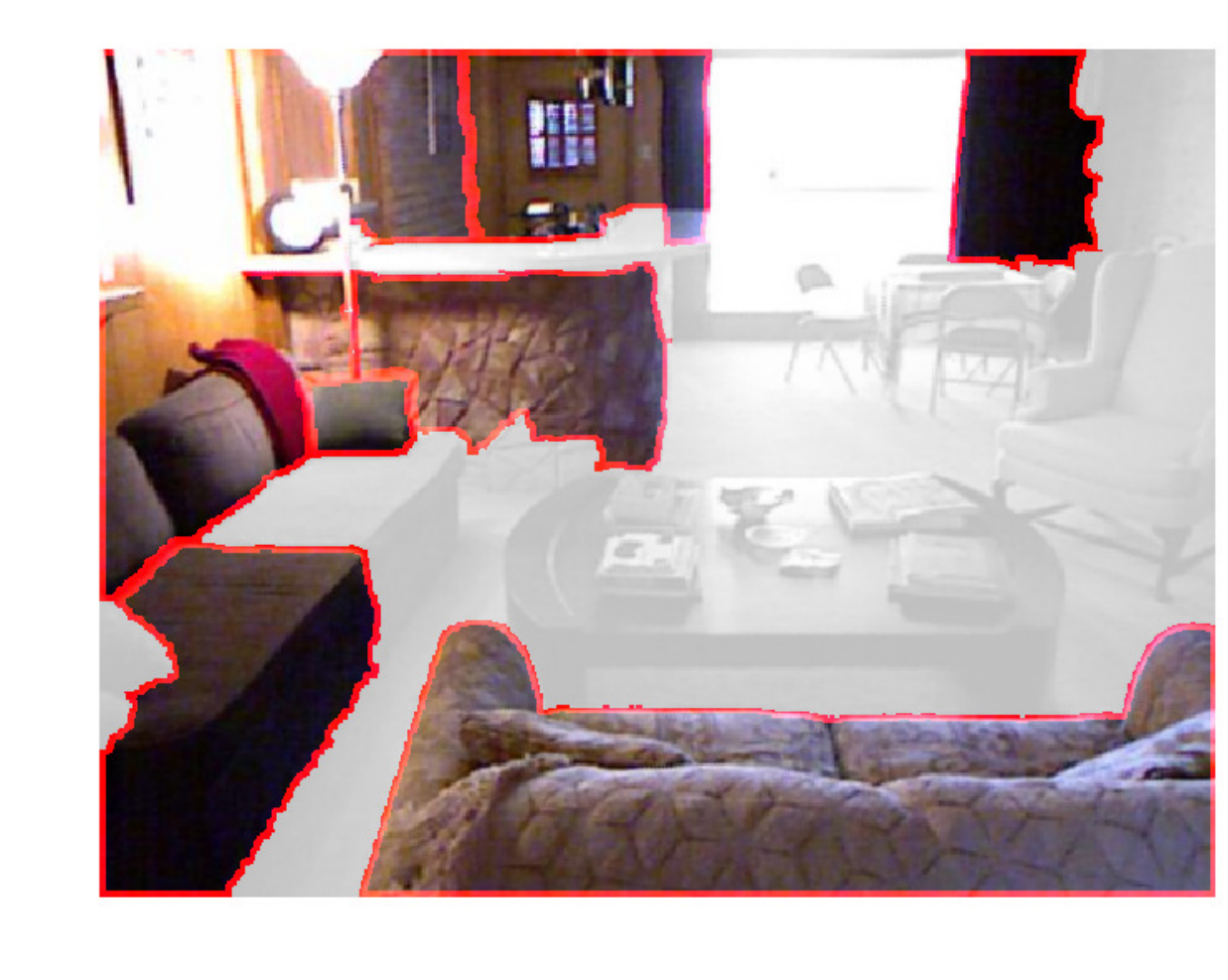}
\includegraphics[width=0.24\linewidth, trim=1cm 1cm 0cm 0.5cm,clip=true]{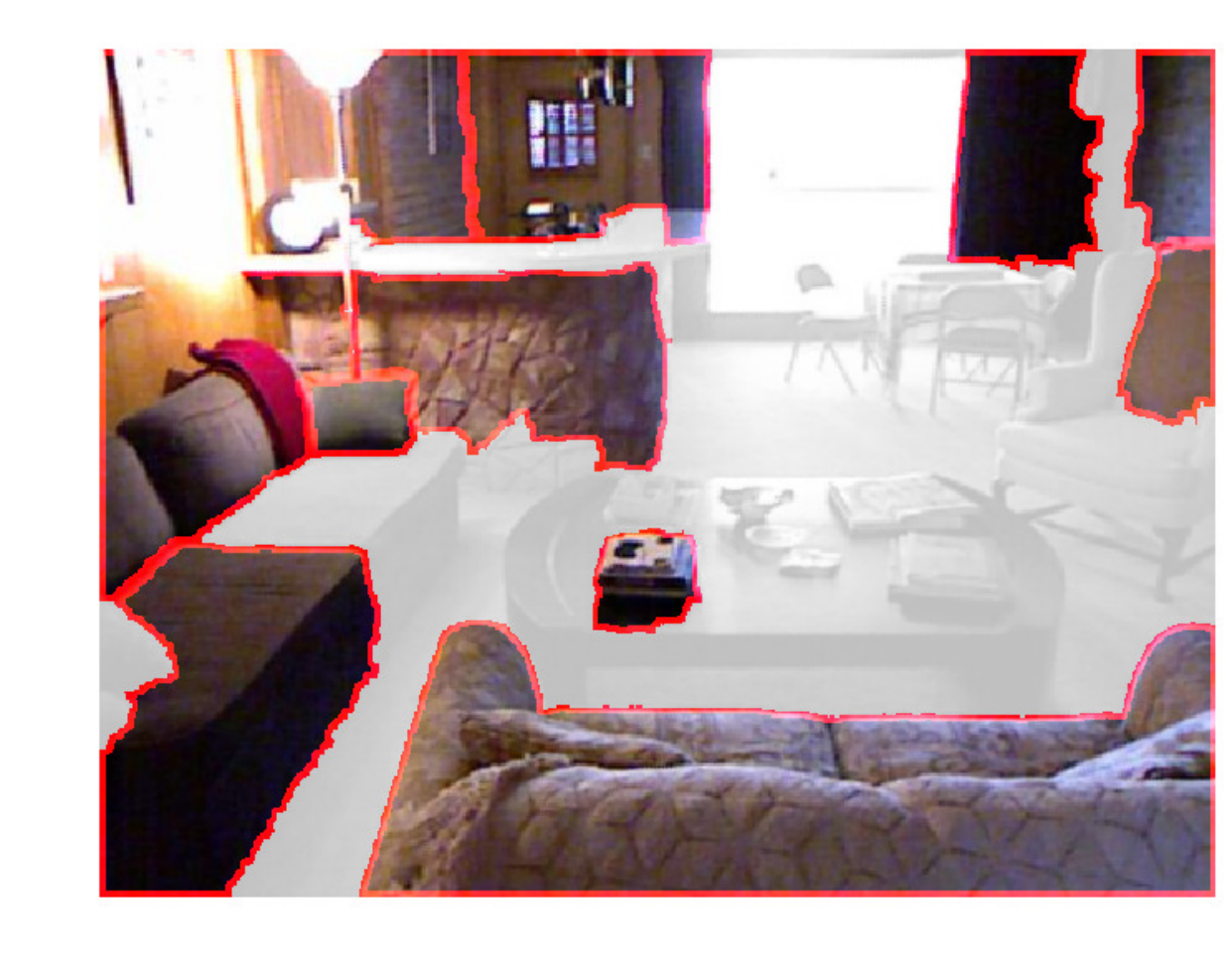}}
\subfigure[Sequence obtained from a search strategy that uses structure in the scene.]{
\includegraphics[width=0.24\linewidth, trim=1cm 1cm 0cm 0.5cm,clip=true]{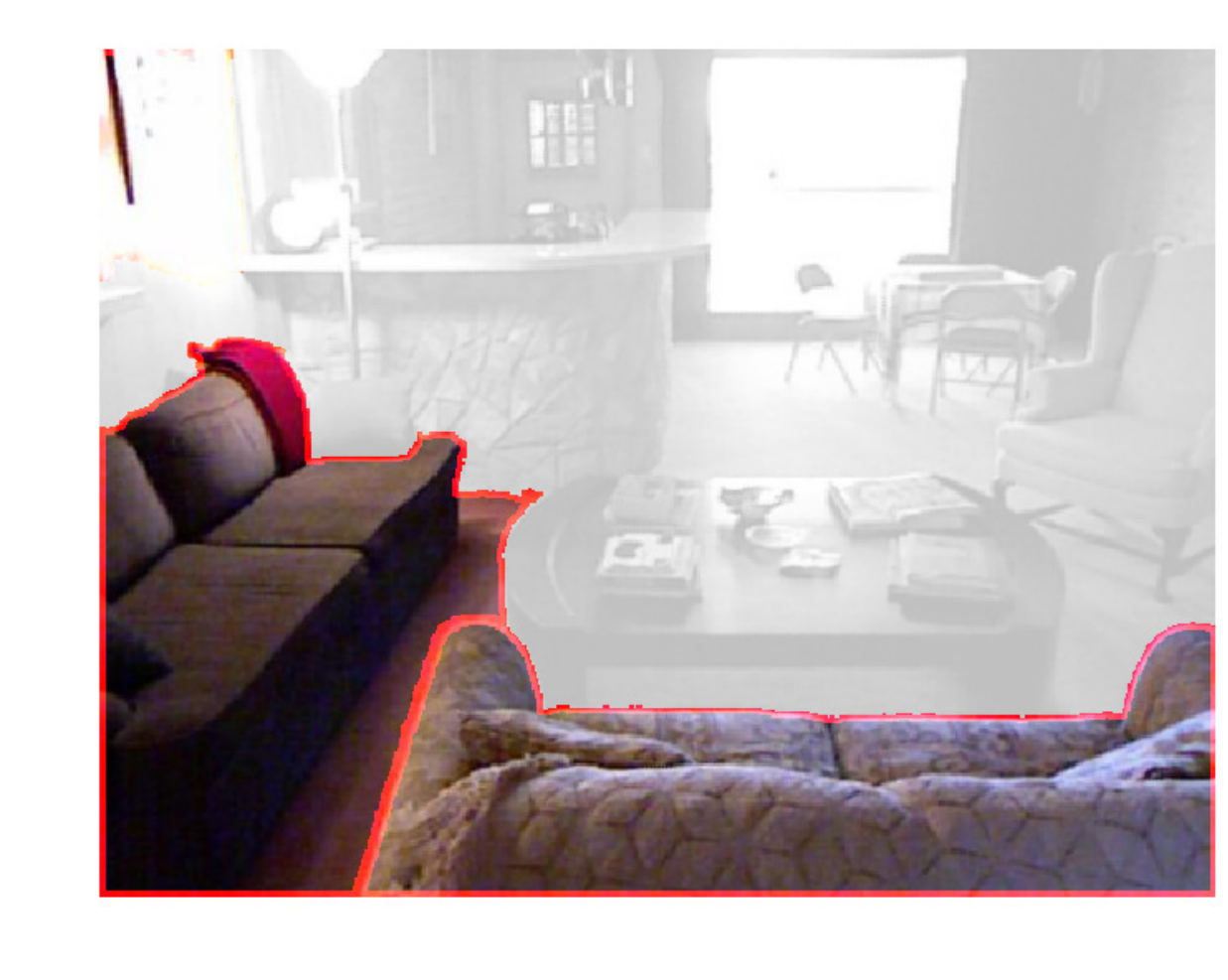}
\includegraphics[width=0.24\linewidth, trim=1cm 1cm 0cm 0.5cm,clip=true]{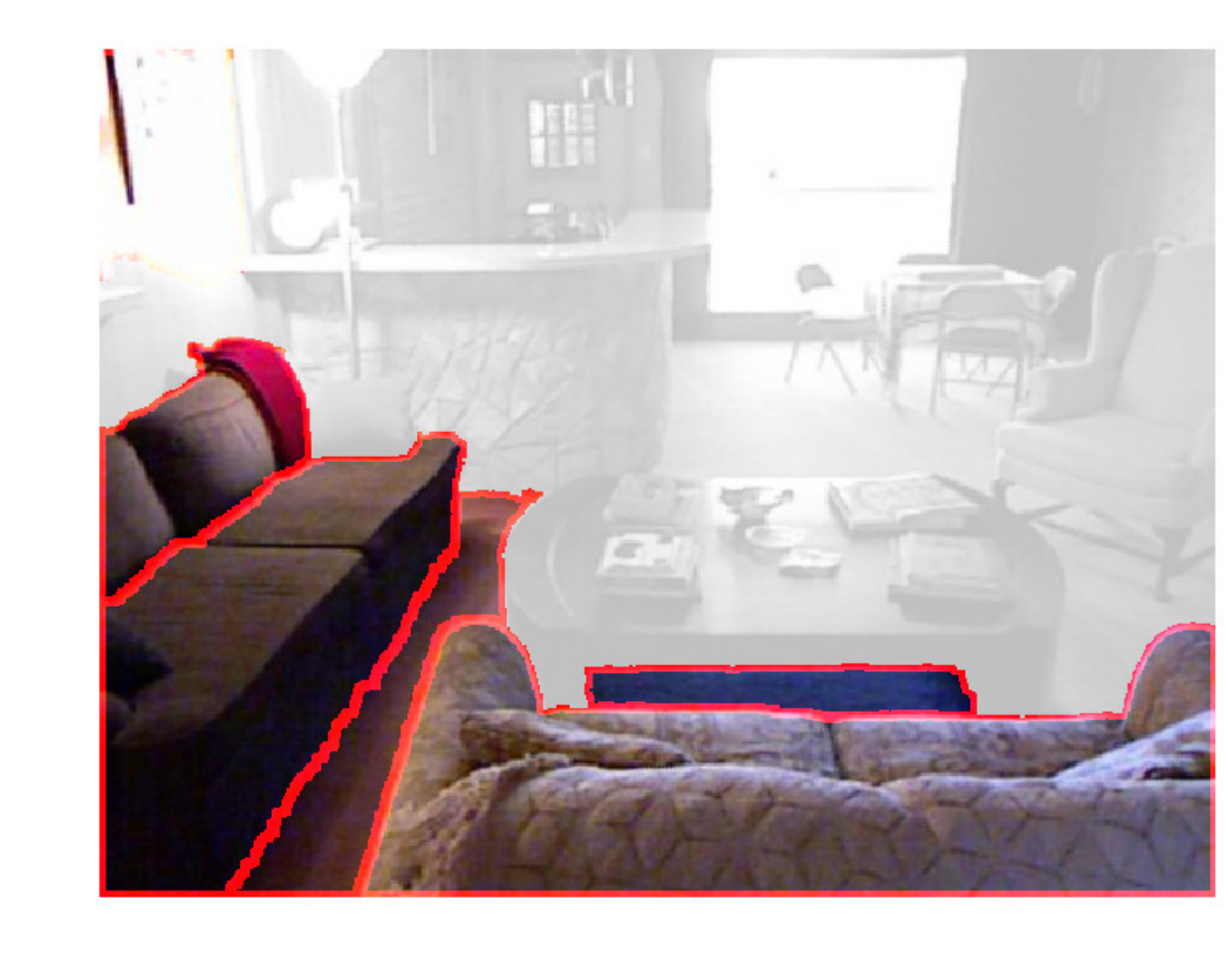}
\includegraphics[width=0.24\linewidth, trim=1cm 1cm 0cm 0.5cm,clip=true]{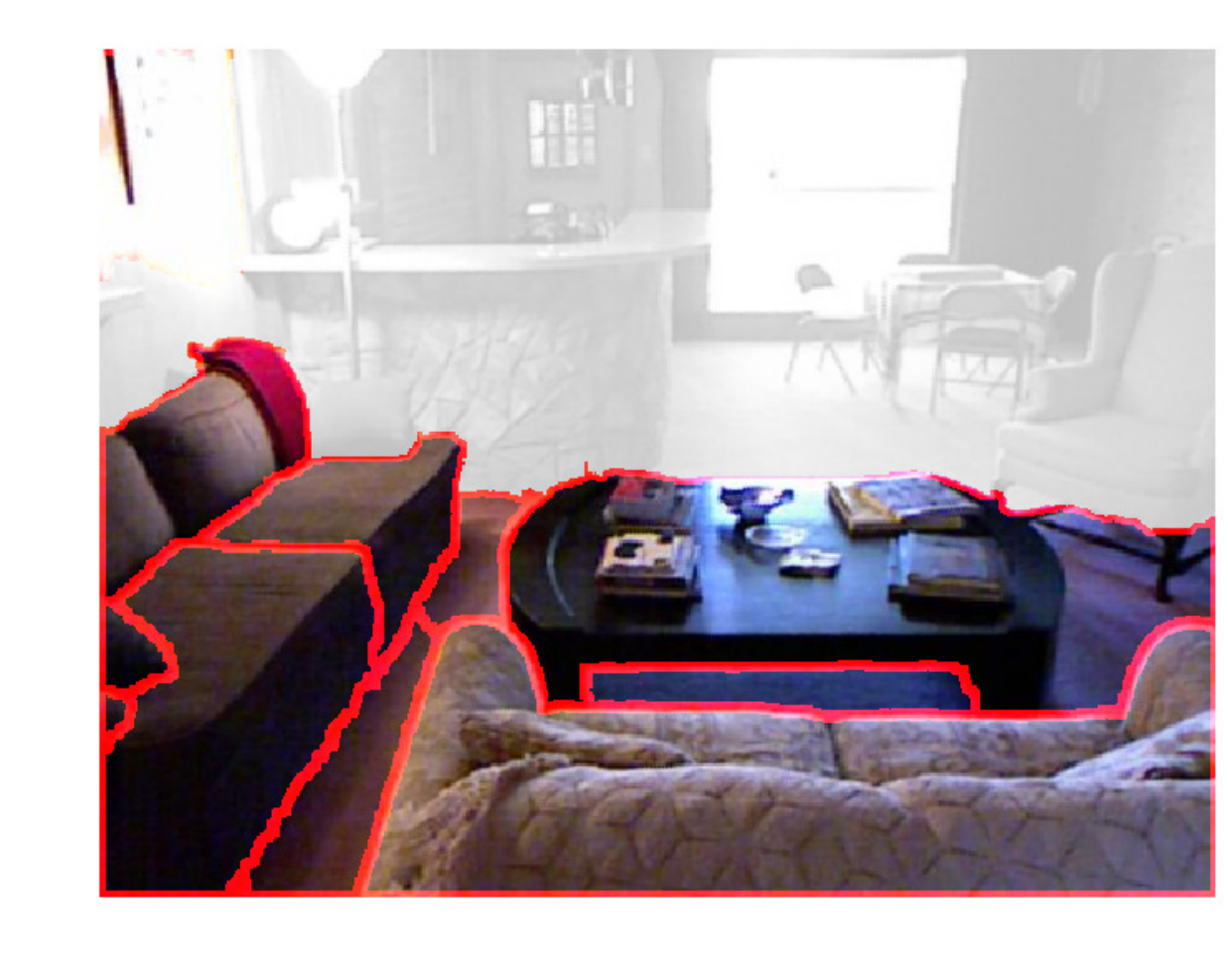}
\includegraphics[width=0.24\linewidth, trim=1cm 1cm 0cm 0.5cm,clip=true]{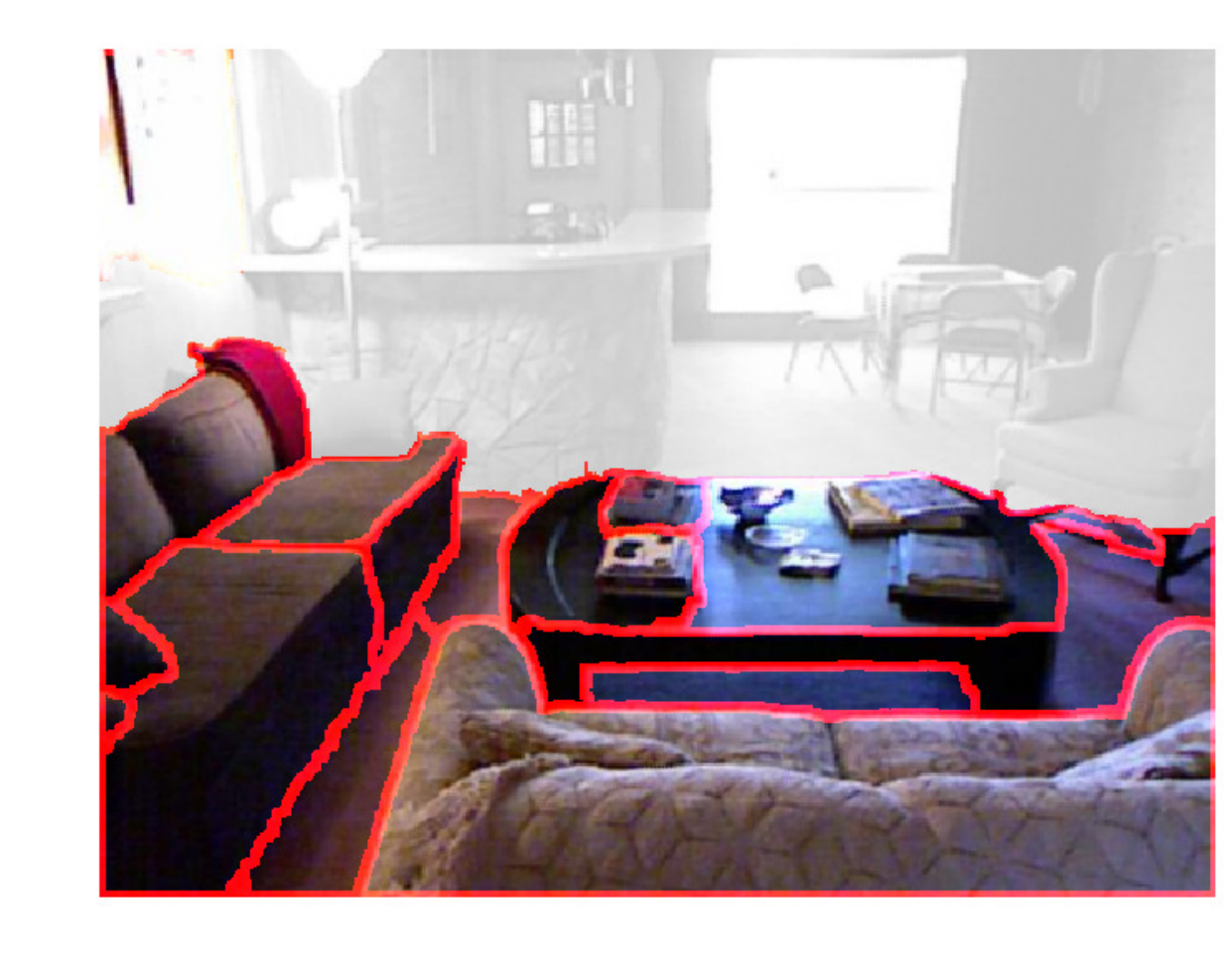}}
\caption{\textbf{Searching for a table}. Each step in the above sequence shows exploration of three additional regions in the image. The search strategy learned using our method utilizes the room structure and the presence of other objects in the image to discover the table region much earlier than using the ranked sequence from an object proposal technique.}
\vspace{-0.15in}
\label{Fig:DemoSequence}
\end{figure*}

Intelligent search strategies can be learned only in domains that contain sufficient structure in the scenes. Frequently recurring patterns between the constituent objects of a scene are essential to learn powerful strategies and predict exploration paths with high confidence. Such structures can be found in indoor scenes of houses, stores and buildings. Hence we illustrate our technique on the indoor scene dataset, NYU depth v2 \cite{Silberman2012}. The other advantage of indoor scenes is the availability of depth data. Gupta \etal \cite{Gupta2014} showed that RCNN \cite{Girshick2014} trained with depth information greatly improved object detection performance. Apart from improving detection performance, depth information provides spatial context that is highly informative for efficient localization of objects. Our experiments show that given a fixed number of regions that can be processed, our sequential exploration technique provides a better average precision than using a ranked sequence provided by the object proposal technique. Figure \ref{Fig:DemoSequence} shows a sequence of regions explored by a strategy trained to detect a table. We compare the search sequence produced by our technique to the ranked sequence provided by the region proposal technique. Our technique is able to utilize the room structure and the presence of other objects in the image to explore the table region much earlier than the object proposal ranking.

\section{Related Work}
Many techniques reduce the number of image windows to limit computation time for object detection. For example, object proposal techniques \cite{Arbelaez2014, Zitnick2014} rank regions in an image based on their likelihood of containing an object. The ranking can be used to prioritize regions for running an object classifier depending on the available computation budget. Such object proposal techniques use only low level image information and do not exploit scene structure.

Some techniques iteratively run the classifier on a few windows and find the next set of windows to be processed based on feedback from the classifier scores and/or spatial context. Lampert \etal \cite{Lampert2008} prune the space of windows using a branch and bound algorithm. Butko and Movellan \cite{Butko2009} use a Partially Observable Markov Decision Process to sequentially place a \textit{digital fovea} (a center of fixation) to detect a single target in an image. Neither of these techniques make use of spatial context between objects to improve window selection. Alexe \etal \cite{Alexe2012} use only spatial context to choose a set of windows to be processed. The classifier is run at the end of window set selection and the maximum scoring window is output as the object location. Gonzalez-Garcia \etal \cite{Gonzalez-garcia} use both spatial context and the classifier scores of previously explored regions. While their output during the testing stage is a sequence of regions, the training is performed without taking into consideration the states (set of objects explored until a step) encountered in a sequence. Hence, they can only model pairwise constraints between an unexplored region and an object. Our technique models relationships between an unexplored region and all the explored objects. Unlike existing work, we use a framework that allows training and testing using the same procedure, thus reducing the burden of tuning many modules in the system.

The idea of sequentially processing an image by exploiting structure is not just relevant to object localization. 
Sequential processing has also been explored for video event detection, where running a multitude of detectors at all spatio-temporal scales is very expensive. Amer \etal \cite{Amer2012} propose an explore-exploit strategy that schedules processes of top-down inference using activity context and bottom-up inference using activity parts. They use a Q-learning algorithm to learn the optimal actions to perform at a state. However, the learning algorithm needs the specification of a reward function which is difficult to obtain in many domains. We use an imitation learning algorithm that alleviates the problem of choosing a reward function.

\section{Sequential Exploration}
\label{Sec:SeqExplore}
The most common formalism for sequential decision making is the Markov Decision Process (MDP). An MDP is characterized by a set of states $S$, a set of actions $A$, transition probabilities $P$ and a reward function $G$ (or equivalently a loss function). A policy $\pi$ is a function that maps states to actions $\pi(s)$. The goal is to find a policy that will maximize a cumulative function of the reward. When the transition probabilities are unknown, reinforcement learning techniques are used to interact with the problem domain and sample the probabilities. 

Our problem is to locate objects of a query class ($q$) by exploring as few image regions as possible. 
Let $R$ be the set of indices of the regions in the image and $t$ correspond to a step index. Let $R_{e}^t$ be the set of indices of the explored image regions and $R_{u}^t = R \setminus R_{e}^t$ be the set of indices of the unexplored image regions at a step $t$. To state our problem in the reinforcement learning setting, a state $s_t$ is the set of all the image regions ($r$) explored until that step. 
\begin{equation}
s_t = \{r_i | i \in R_e^t\}
\end{equation}

An action corresponds to selecting the next image region to explore, $a_t = r_j$  where $j \in R_{u}^t$.  
The reward function is difficult to specify for our image exploration problem.
If we assume that spatial arrangements of objects in images are generated from a hidden distribution similar to games generated by a hidden emulator, a true reward function will allow the policy to learn a predictor that can replicate the behavior of the hidden distribution. For example, if we are searching for a chair, by setting the reward values higher for regions near a table than those far away from it, the policy assigns a greater importance to table proximity feature. Since images contain samples of spatial arrangements from the hidden distribution and the true reward values are unknown, an imitation learning algorithm can be used to learn the optimal policy. 
In imitation learning \cite{Abbeel2004}, rather than specifying a reward function, an oracle demonstrates the action to take and the policy learns to imitate the oracle. For us, the oracle selects the next region to explore based on the groundtruth labels. Hence, the policy is trained to predict labels similar to the groundtruth.

Imitation learning algorithms usually learn a strategy by training a classifier on the dataset of state features and actions (labels) obtained by sampling sequences produced by an oracle policy. They make an i.i.d assumption about the states encountered during the execution of a learned policy which does not hold for our problem since the policy's prediction affects future states. During the test stage, if the policy encounters a state that was not generated by the oracle policy, it could predict an incorrect action that can lead to compounding of errors. Ross \etal \cite{Ross2011} propose an imitation learning algorithm called DAgger (Dataset Aggregation) that does not make the i.i.d assumption about the states. DAgger finds a policy $\hat{\pi}$ which minimizes the observed surrogate loss $\ell(s,\pi)$ under its induced distribution of states,
\begin{equation}
\hat{\pi} = \argmin_{\pi \in \Pi}  \mathbb{E}_{s\sim d_{\pi}}[\ell(s,\pi)]
\end{equation}
where $d_{\pi} = \frac{1}{T}\sum_{t=1}^T d_{\pi}^t$ is the average distribution of states if we follow policy $\pi$ for $T$ steps. Since $d_{\pi}$ is dependent on the policy $\pi$, this is a non-i.i.d supervised learning problem. 

In our work, $\ell(s,\pi)$ is the Hamming loss of $\pi$ with respect to $\pi^*$ - the oracle policy.  At a given state, the policy is penalized for what it predicts for all the regions in the unexplored set. Let 
\begin{equation}
\bm{p} = (p_i)_{i \in R_u^t}
\end{equation}
be a list of the predicted labels where each label $p_i \in \{0,1\}$ indicates whether the corresponding region is predicted to contain an object of the queried class or not. The oracle policy produces a list the same length as $\bm{p}$ with groundtruth labels. The Hamming loss is measured between the list of predicted labels and groundtruth labels. When the policy labels more than one region for exploration, we select the region with the highest belief as the next region to explore.

DAgger trains a single cost sensitive classifier for policy $\hat{\pi}$ that considers features extracted from a state and predicts labels to determine the next action. During training, it starts with an initial classifier and runs through the states, predicting labels for each state. Based on its predictions, it is assigned a loss value at each state. At the end of an iteration, all the features, the predicted labels and the loss values for all states are collected. The aggregate of all the collected datasets until the current iteration is used to train a cost sensitive classifier, which becomes the policy for the next iteration. DAgger is available through a simple interface in the Vowpal Wabbit\footnote{\url{https://github.com/JohnLangford/vowpal_wabbit}} library. It contains a new programming abstraction proposed by Daum\'{e} III \etal \cite{DaumeIII2014} where a developer writes a single \textit{predict} function that encodes the algorithm for the testing stage and the training is done by making repeated calls to this predict function.

\begin{algorithm}[t!]
\footnotesize
\caption{Sequential Exploration}
\begin{multicols}{2}
\begin{algorithmic}[1]
\Function{seq\_explore}{$obj\_proposals, N$}
    \State $explored\_list \gets \varnothing$
    \State $curr\_regions \gets obj\_proposals[0]$
    \State $unexplored\_regions \gets obj\_proposals[1:\text{end}]$
    \State $i \gets 0$
    \While{$i < N$ and $curr\_regions \neq \varnothing$}
        \State $i \gets i+1$
    	\State $r_{curr} \gets \Call{pop}{curr\_regions}$
   	\State \Call{Push}{$explored\_list$, $r_{curr}$}
	\For{$r_j \in unexplored\_regions$}
		\State $score_j = \Call{classify}{r_j,explored\_list}$
	\EndFor
	\State $next \gets \argmax_j score_j$
	\State \Call{push}{$curr\_regions, r_{next}$}
	\State \Call{remove}{$unexplored\_regions, r_{next}$}
    \EndWhile
    \State \Return{$explored\_list$}
\EndFunction
\end{algorithmic}
\columnbreak
\begin{algorithmic}[1]
\Function{classify}{$r_j,explored\_list$}
    \State $features \gets \varnothing$
    \State $unary_j \gets \Call{unary\_features}{r_j}$
    \State \Call{append}{$features, unary_j$}
    \State \Call{non\_maximal\_supress}{$explored\_list$}
    \State $pairs \gets \varnothing$
    \For {$r_k \in explored\_list$}
    	\State $label \gets \Call{query\_label}{r_j}$
	\If{$label \neq \text{bgnd}$}
   	    \State $pair_{r_j, r_k} \gets \Call{pair\_features}{r_j, r_k}$
    	    \State \Call{append}{$pairs, pair_{r_j, r_k}$}
	\EndIf
    \EndFor
    \State $agg\_pair\_features \gets \Call{agg\_stats}{pairs}$
    \State \Call{append}{$features, agg\_pair\_features$}
    \State $(label,score) \gets \Call{DAgger\_Predict}{features}$
    \If{training}
        \State \Call{DAgger\_Setloss}{$label,\text{groundtruth}$}
    \EndIf
    \State \Return $score$
\EndFunction
\end{algorithmic}
\end{multicols}
\label{Alg:SeqExplore}
\vspace{-0.15in}
\end{algorithm}

The function SEQ\_EXPLORE shown in Algorithm \ref{Alg:SeqExplore} is substituted for the \textit{predict} function in the programming abstraction of Daum\'{e} III \etal \cite{DaumeIII2014}. The input to the algorithm is a list of object proposals and the number of regions that we are allowed to process. We use a modified MCG \cite{Arbelaez2014, Gupta2014} for region proposal generation and RCNN-depth \cite{Gupta2014} for region classification. 
The unary features used for classification are objectness score, proposal rank, mean depth of the region, mean distance from the back of the room, minimum height from the ground and maximum height from the ground. The pairwise features are 2D area overlap, 2D size ratio, distance between centroids, difference in mean distance from the back of the room, difference in minimum heights from the ground and the difference in maximum heights from the ground. Most of these features were used by Silberman \etal \cite{Silberman2012} for performing support inference. The aggregate feature set is constructed by performing min-pooling for each class and each pairwise feature. For example, one of the aggregate features would be constructed by collecting all the distances between centroids of the current region and the regions of table class, and then taking the minimum of those distances. This feature measures "how far is the closest table (and every other class) from the current region?"

The computational complexity of our algorithm in the worst case scenario of exploring all regions is $O(n^2)$ where we perform classification for every unexplored region at every step after adding one region. However, we do not repeat the classification if a newly explored region is marked as background since it does not change the context features at that step. Hence the number of iterations where we classify the unexplored regions is dependent on the number of foreground regions ($k$) in the image and the complexity is $O(nk)$. Since there are very few foreground regions in an image, $k$ is usually small. 

\subsection{Data subset selection}
Due to the presence of a large number of background regions, the training process can become very slow. Hence we need to select a subset of the background regions such that the training time becomes tractable while maintaining performance. A popular approach to background set collection is hard negative mining, an iterative process where the training data is progressively augmented with the false positive examples produced by the classifier in an iteration. Hard negative mining is a computationally expensive technique which is exacerbated in our case by the already expensive training process for a search strategy. Instead we use a data subset selection technique motivated by the theory of Optimal Experiment Design (OED) \cite{Pukelsheim}. Given a linear regression model, the goal of OED is to select samples such that the variance in the regression coefficients is minimized. A smaller variance in the coefficients indicates that the prediction error on the test set is low and hence the linear regression model trained with such a subset does not overfit the training data. Since DAgger uses a linear classifier to predict actions, we employ an OED criterion to select a subset of the background samples. 

Let $X$ be a matrix of $n$ samples with $p$ features. Let $\Pi$ be a row selection matrix of size $k \times n$. Each row of $\Pi$ contains a value of one in exactly one column and zeros otherwise.  Let $Y$ be the vector of predicted labels. A linear regression model can be written as
\begin{equation}
Y_{k \times 1} = \Pi_{k \times n} X_{n \times p}\beta_{p \times 1} + \epsilon_{k \times 1}
\end{equation}
$\epsilon$ is the noise vector with mean zero and variance $\sigma^2 I_k$. In ordinary least squares regression, the prediction error is directly proportional to the variance of the regression coefficients. The variance is given by
\begin{equation}
var(\hat{\beta}) = \sigma^2(X^T \Pi^T \Pi X)^{-1}
\end{equation}
Optimal Experiment Design suggests many criteria that optimize the eigenvalues of the inverse covariance matrix as a way to minimize the variance in the regression coefficients. The A-optimal criterion minimizes the trace of the inverse covariance matrix and the D-optimal criterion minimizes the determinant of the inverse covariance matrix. The D-optimal criterion \cite {John1975} is more popular due to the availability of off-the-shelf implementations and also, it simplifies the determinant minimization of an inverse to maximizing the determinant of the covariance matrix. Since we want to select only a subset of the negative samples, we fix the selection variables for the positive samples. We use a row exchange algorithm\footnote{The row exchange algorithm is available as part of the Statistics Toolbox in MATLAB.} that iteratively adds and removes rows based on the increments in the determinant. The features we use in the data matrix for subset selection are only the mean centered unary features, since the pairwise features are constructed dynamically and they are difficult to know beforehand.

\section{Experiments and Results}
\subsection{Dataset}
We demonstrate our approach on the NYU depth v2 dataset \cite{Silberman2012}. We use the RCNN-Depth module of Gupta \etal \cite{Gupta2014} for the region classification. Their region proposal module is a modified Multiscale Combinatorial Grouping (MCG) \cite{Arbelaez2014} technique that incorporates depth features. Their feature extraction module is RCNN \cite{Girshick2014} which includes CNNs fine-tuned on the depth images. The dataset is split into three partitions - 381 images for training, 414 images for validation and 654 images for testing. Since RCNN is trained on the training split, the performance of the detectors on the training set images is extremely good and does not reflect the behavior of the detectors on the test set. Hence we run the detectors on the validation set, obtain groundtruth labels for the detections and this set forms the training set for learning search strategies. The thresholds for the detectors are set based on the best F1 point on the validation set PR curves. We work with 18 categories and do not include the box category as its performance values are very low with an average precision of 1.4\%. The operating characteristics of RCNN-depth can be found in the supplementary material. We consider the top 100 regions obtained from the region proposal module. One of the reasons we use only 100 regions is that as we increase the number of regions, the amount of variation in the background regions increases, making the classification boundaries highly nonlinear given our feature set. Since the number of available positive samples is not sufficiently large, it is difficult to train a nonlinear classifier. 

\subsection{Sequential Exploration}
\begin{figure*}[t!]
\centering
\includegraphics[width=1.2in]{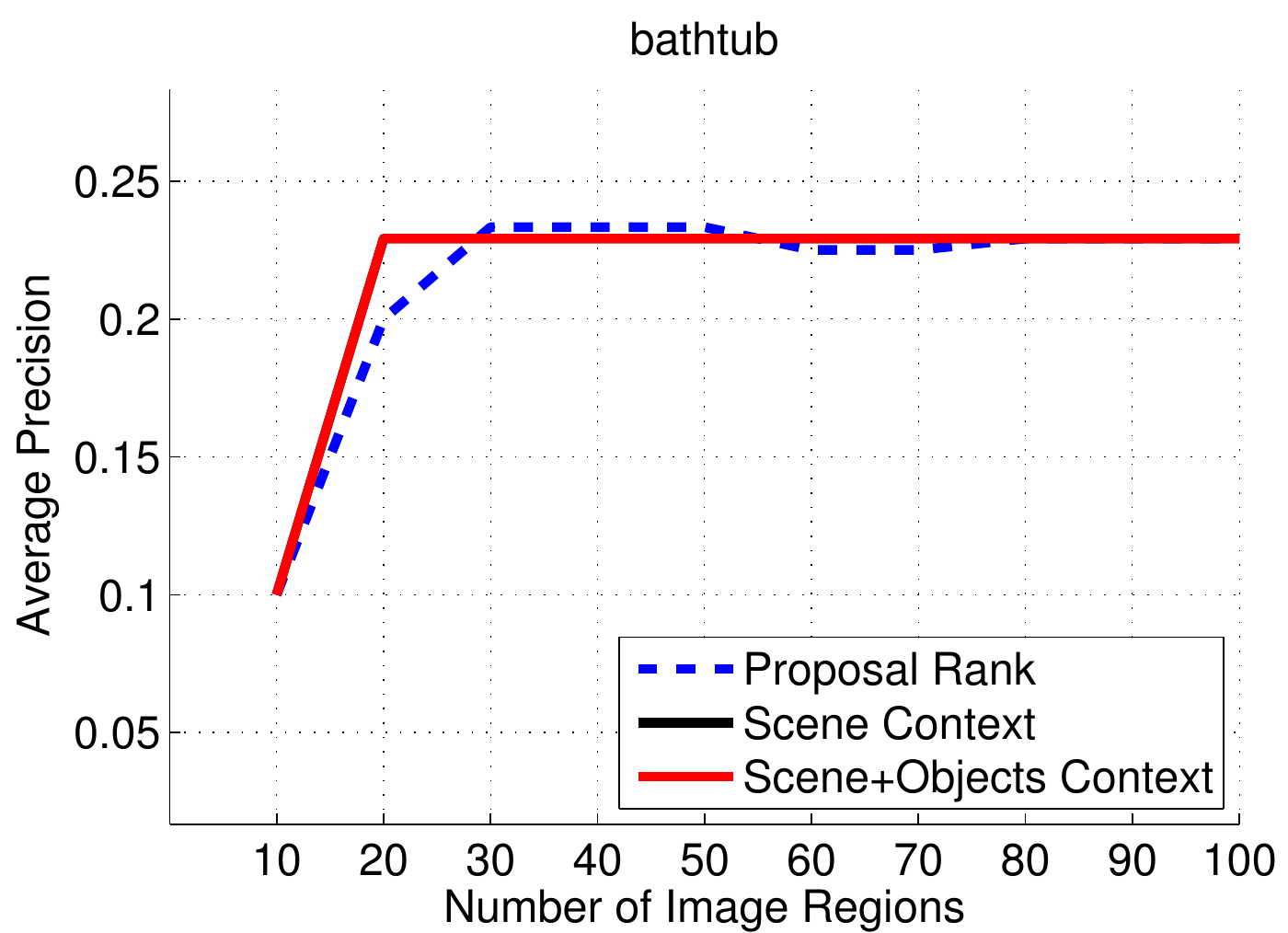} 
\includegraphics[width=1.2in]{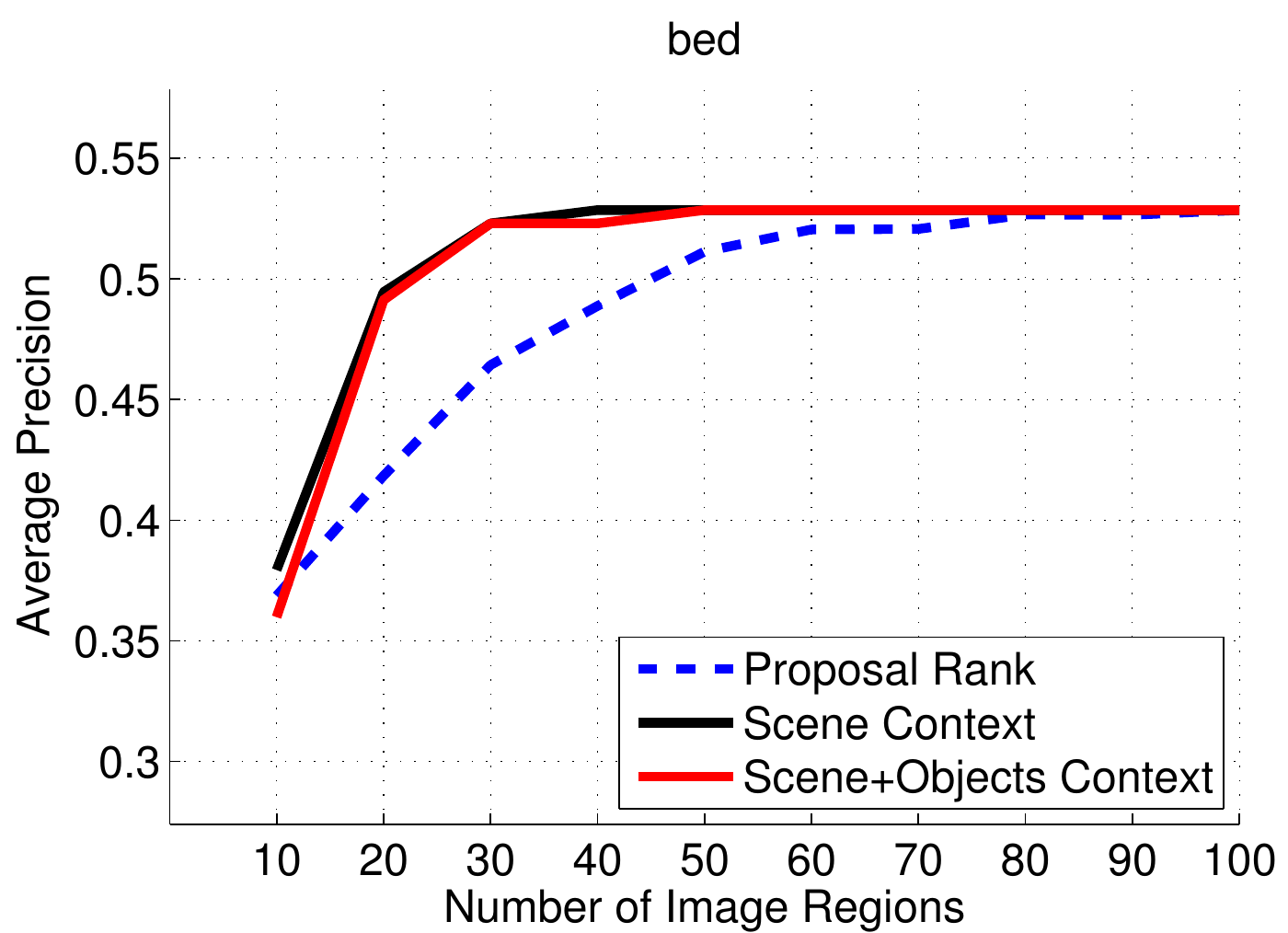} 
\includegraphics[width=1.2in]{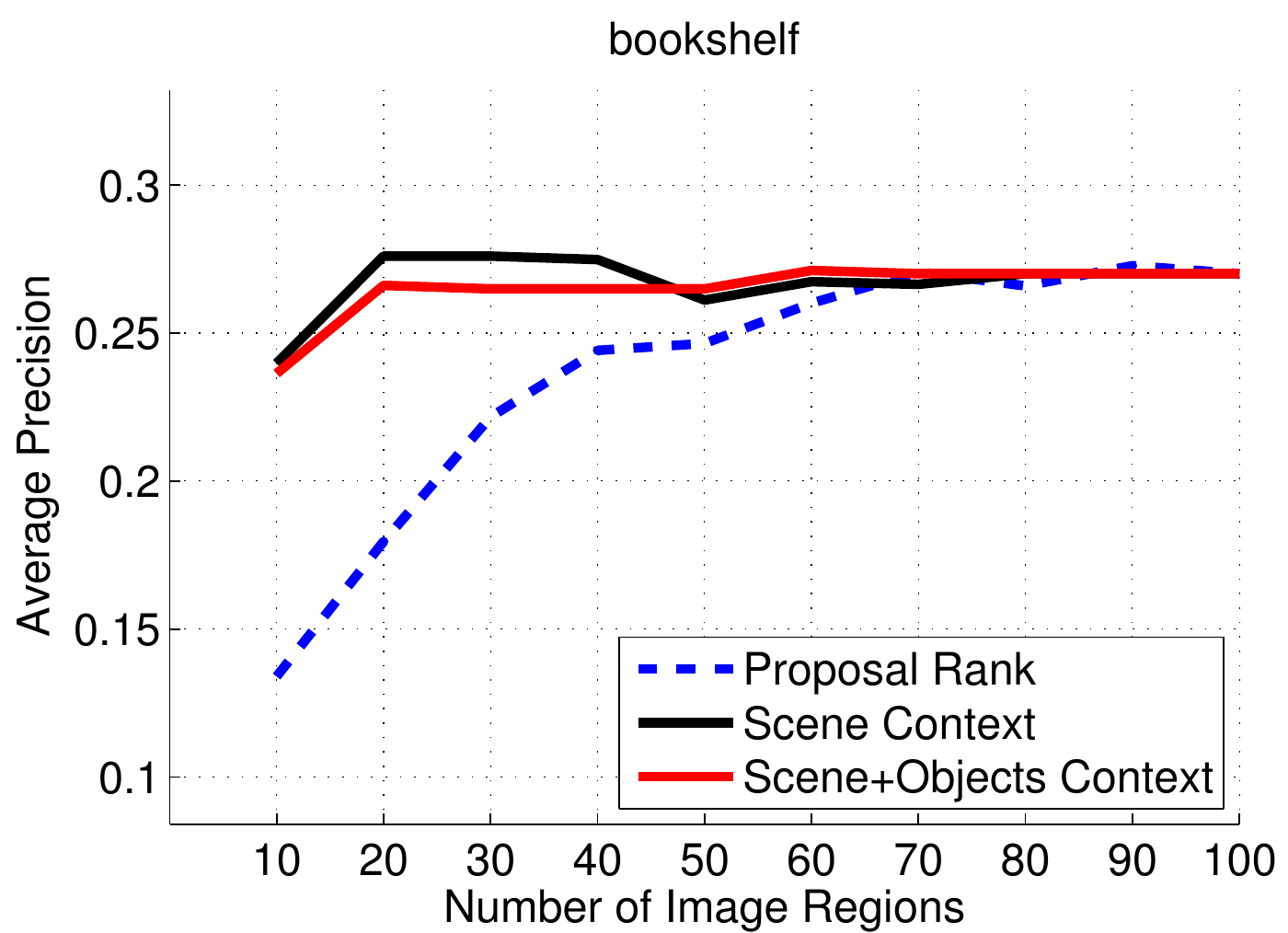} 
\includegraphics[width=1.2in]{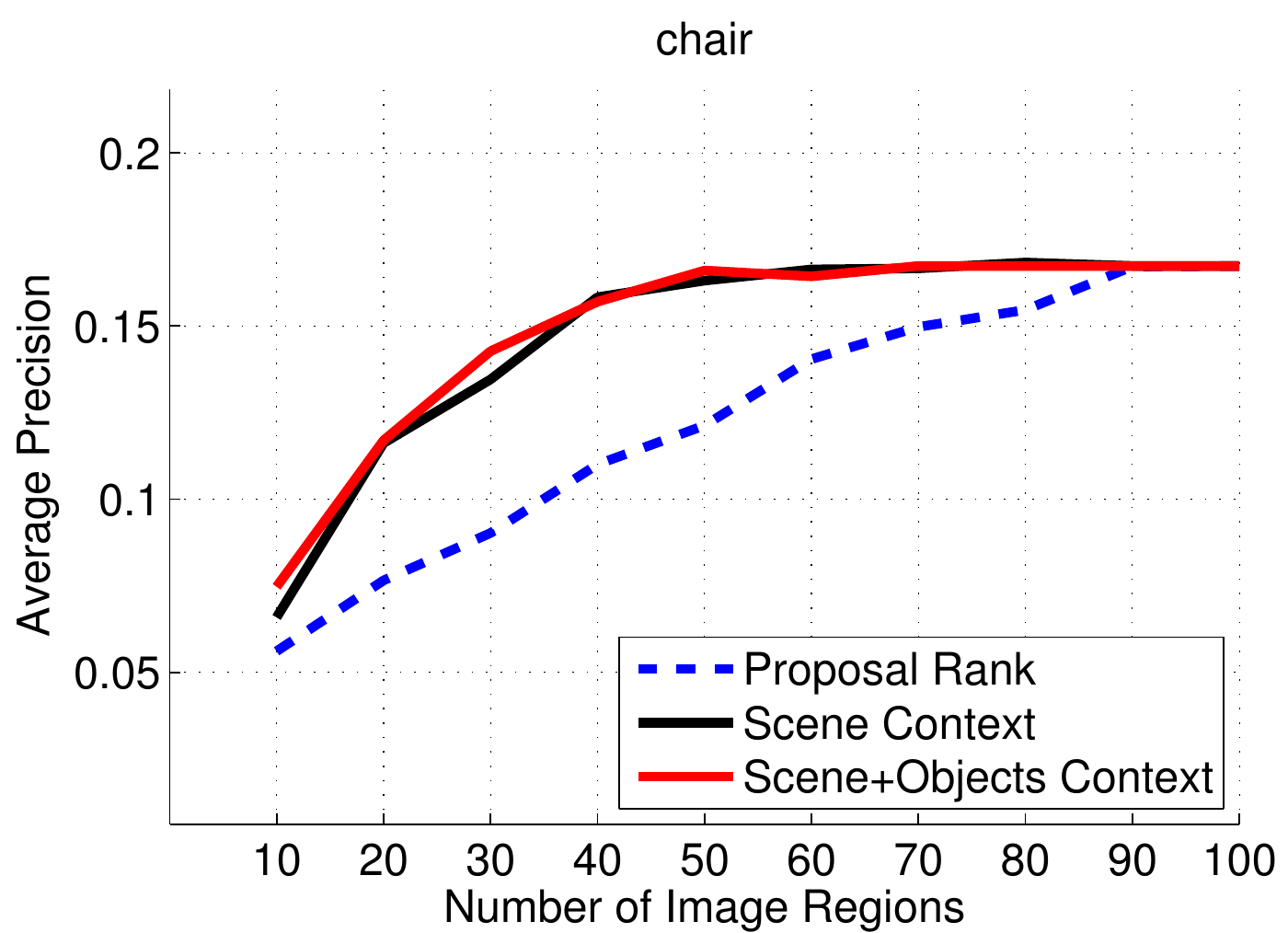} 
\includegraphics[width=1.2in]{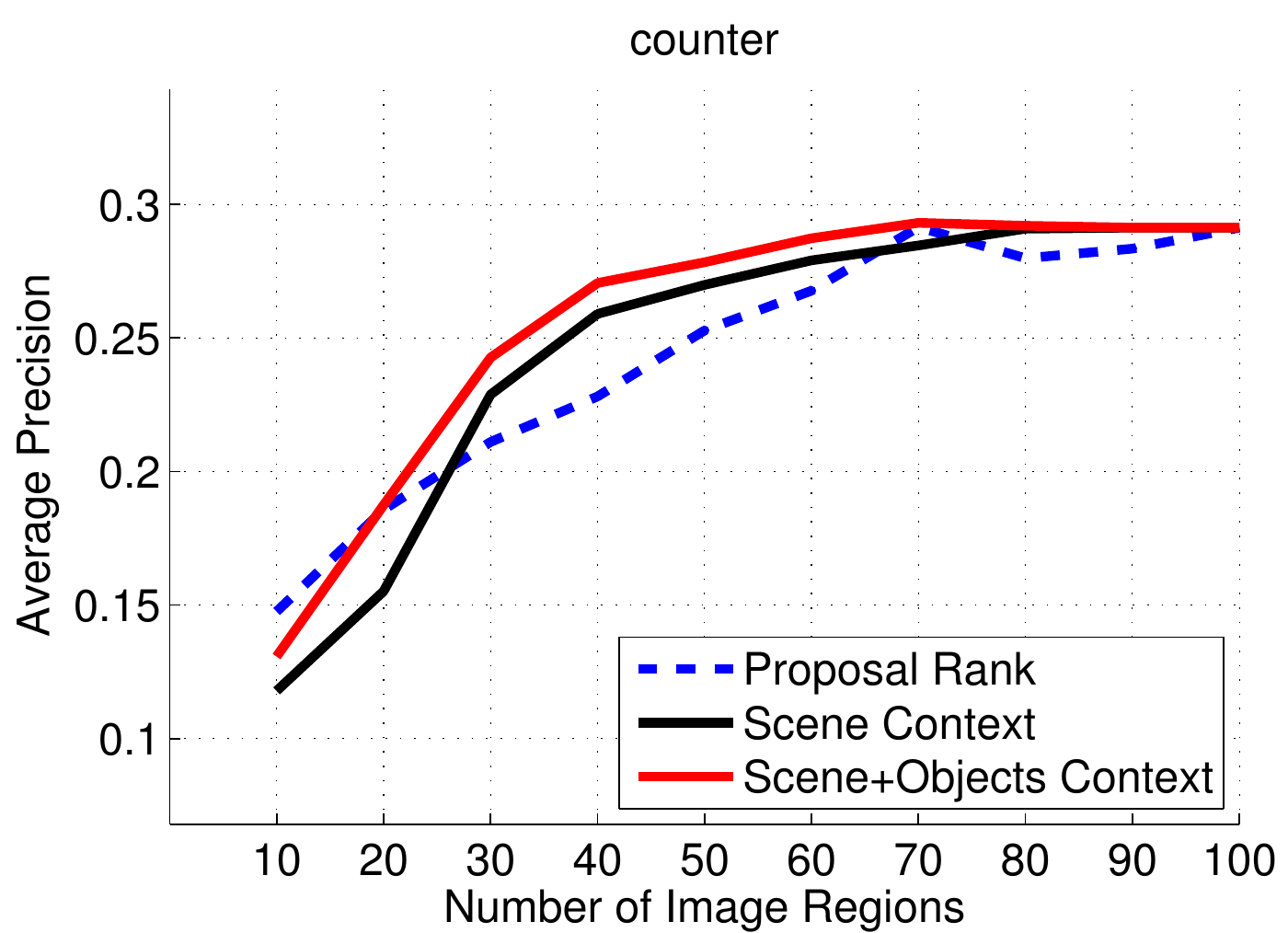} 
\includegraphics[width=1.2in]{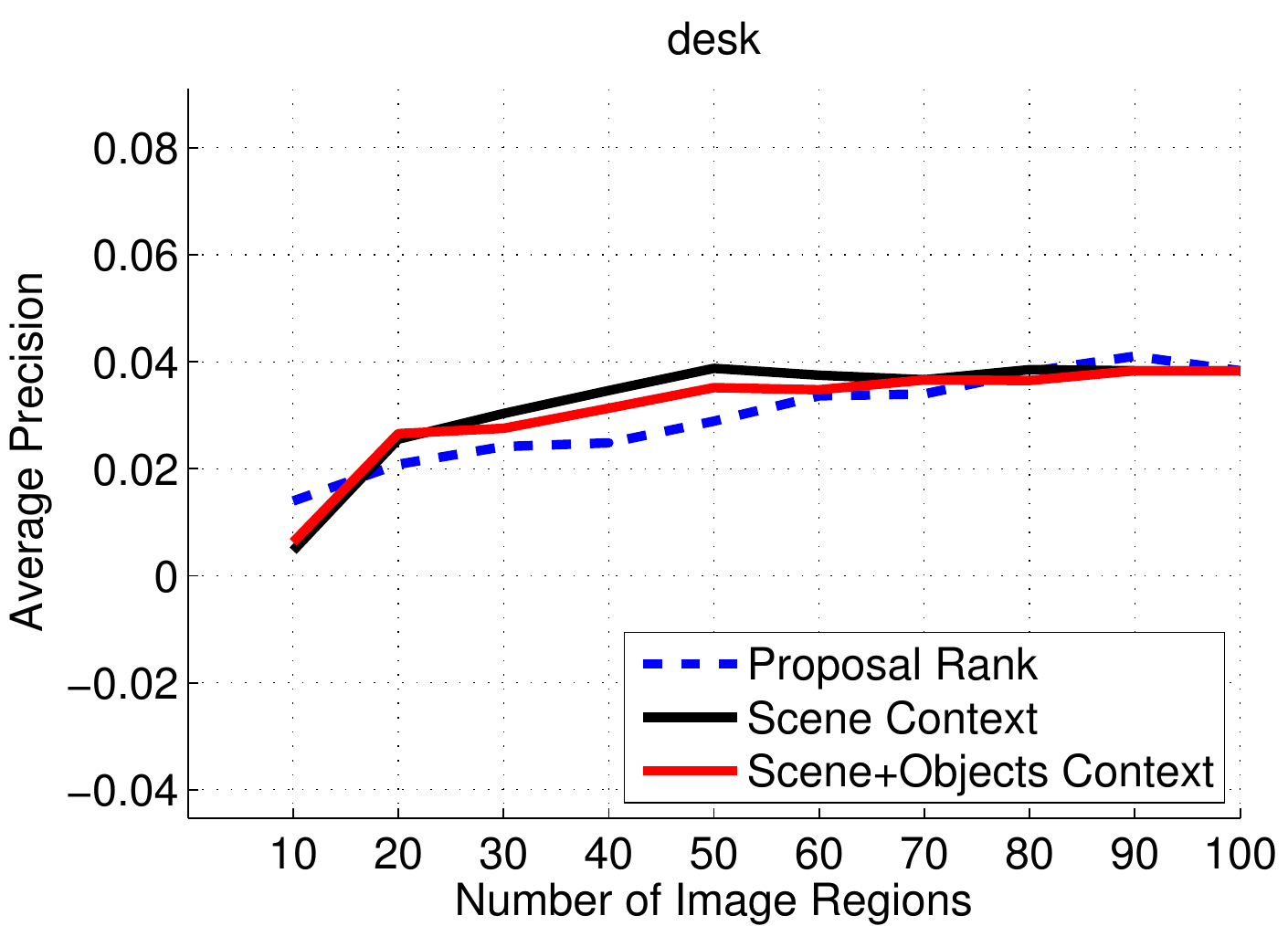} 
\includegraphics[width=1.2in]{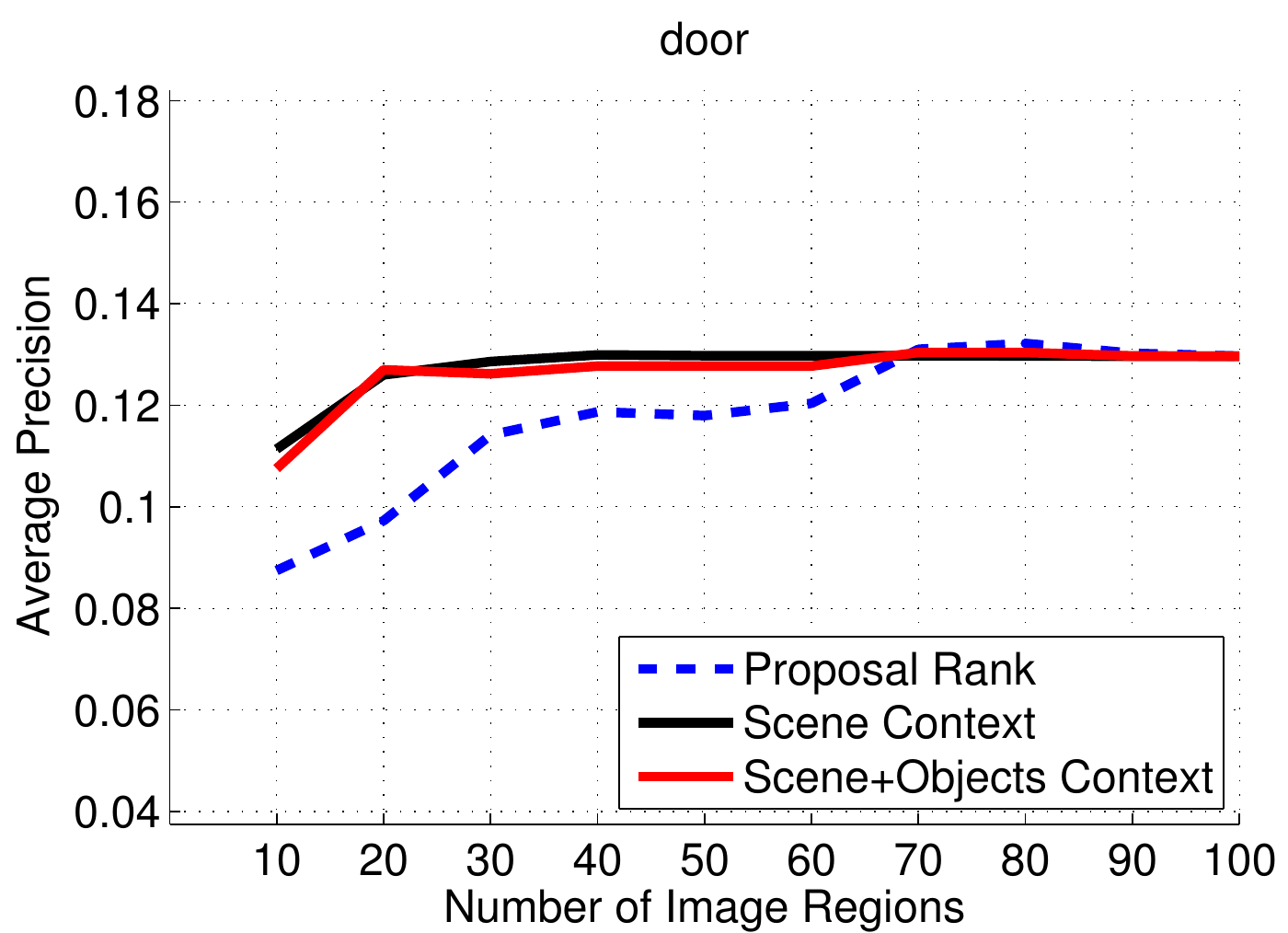} 
\includegraphics[width=1.2in]{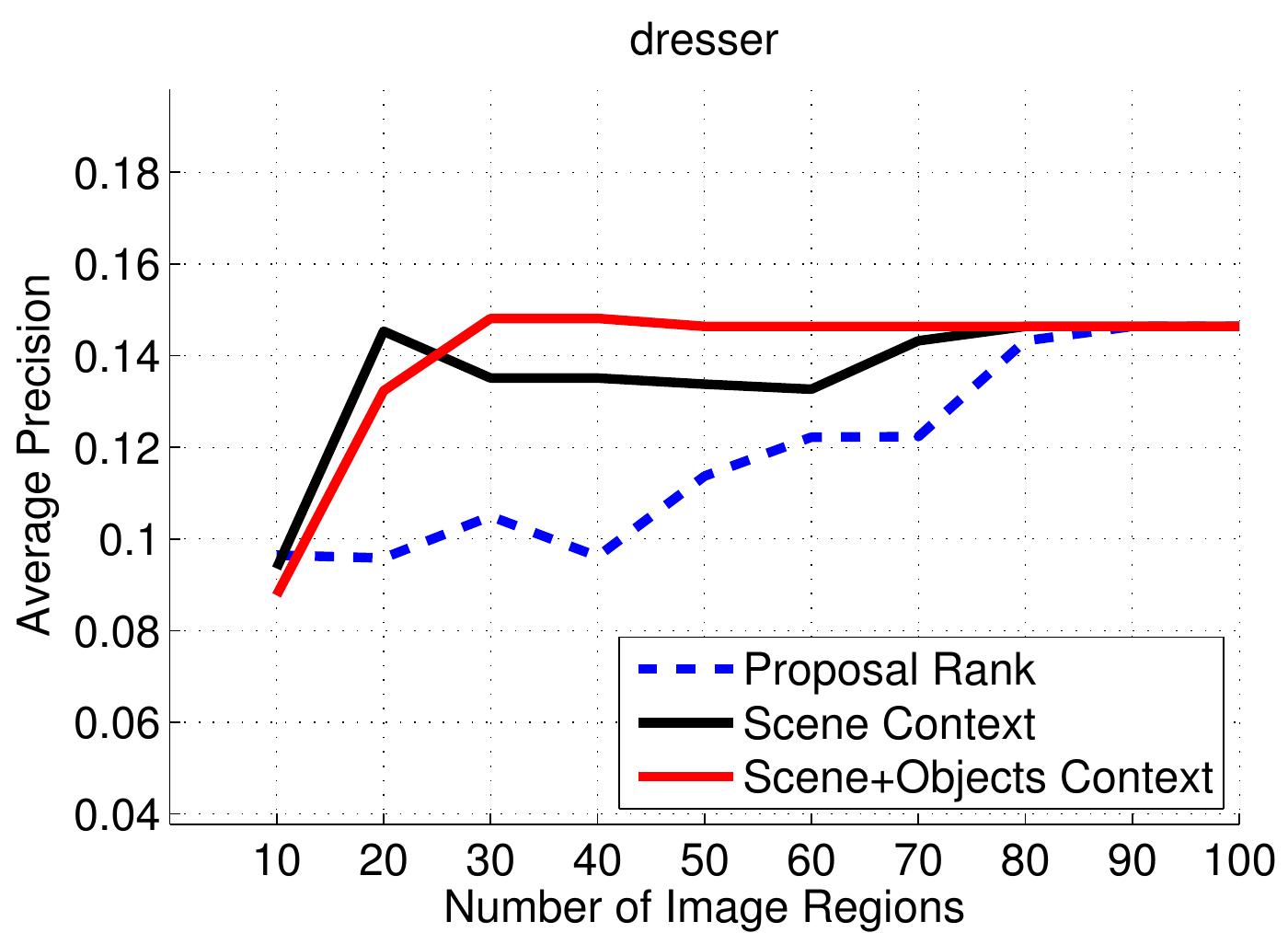} 
\includegraphics[width=1.2in]{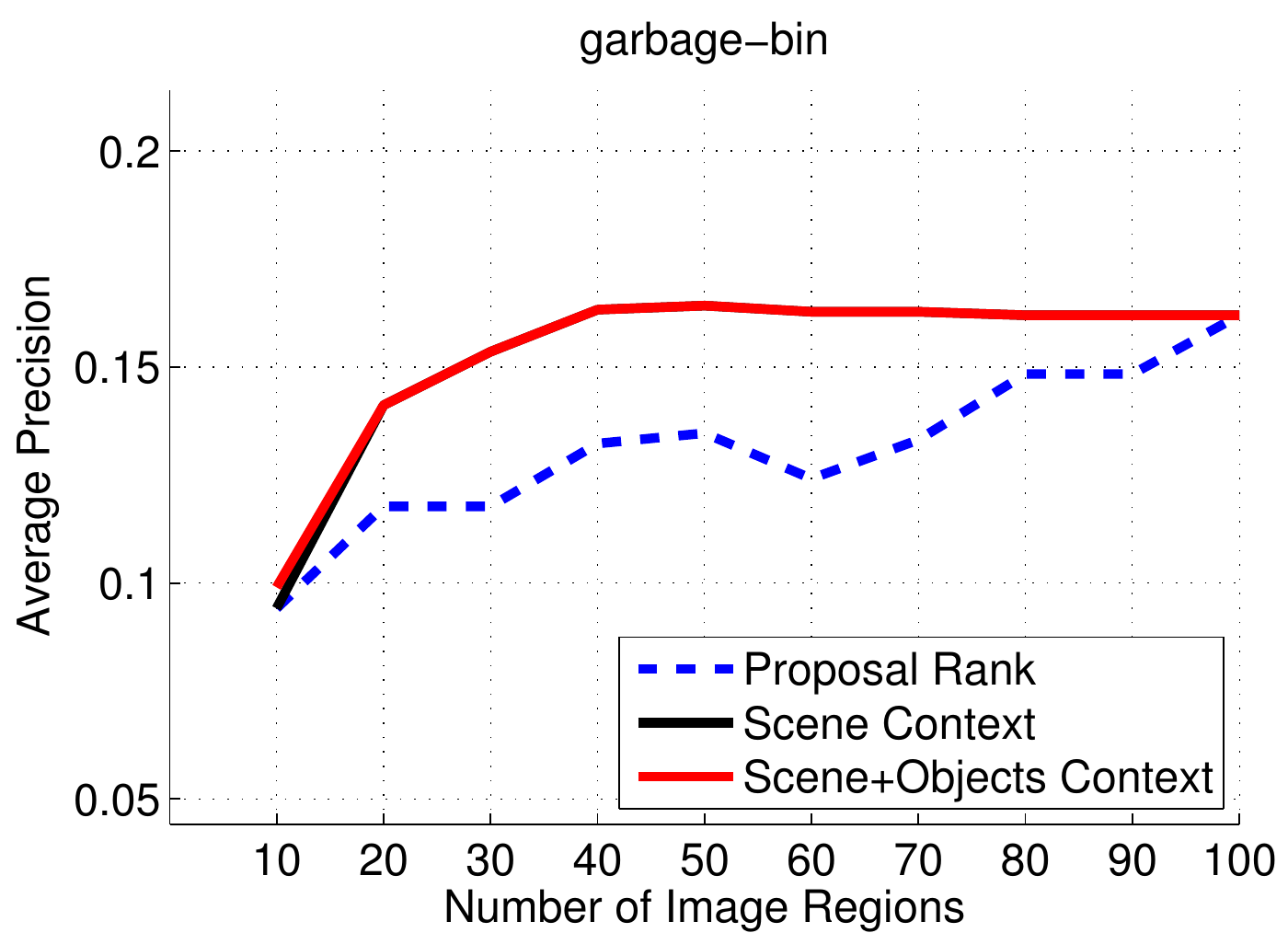} 
\includegraphics[width=1.2in]{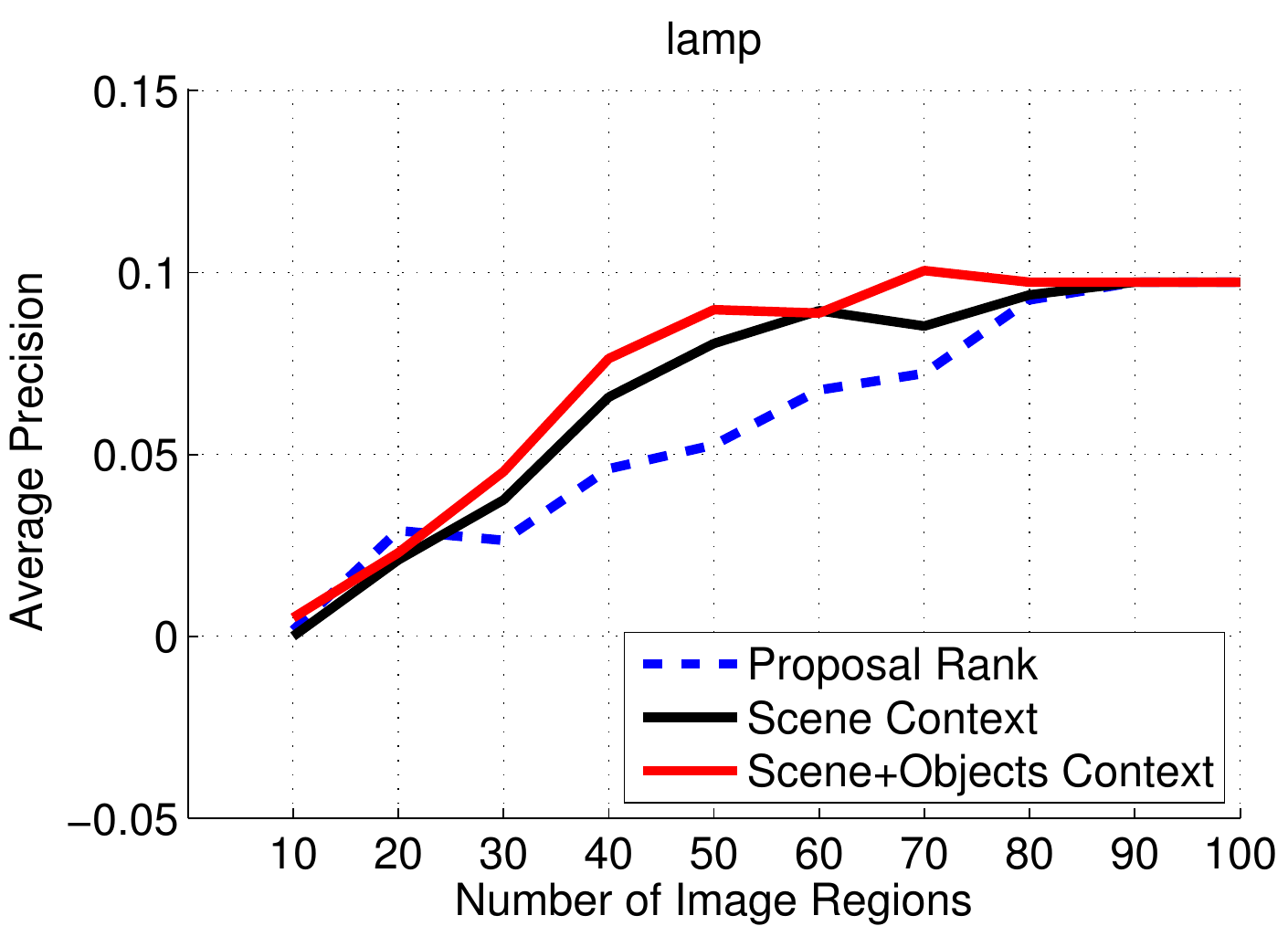} 
\includegraphics[width=1.2in]{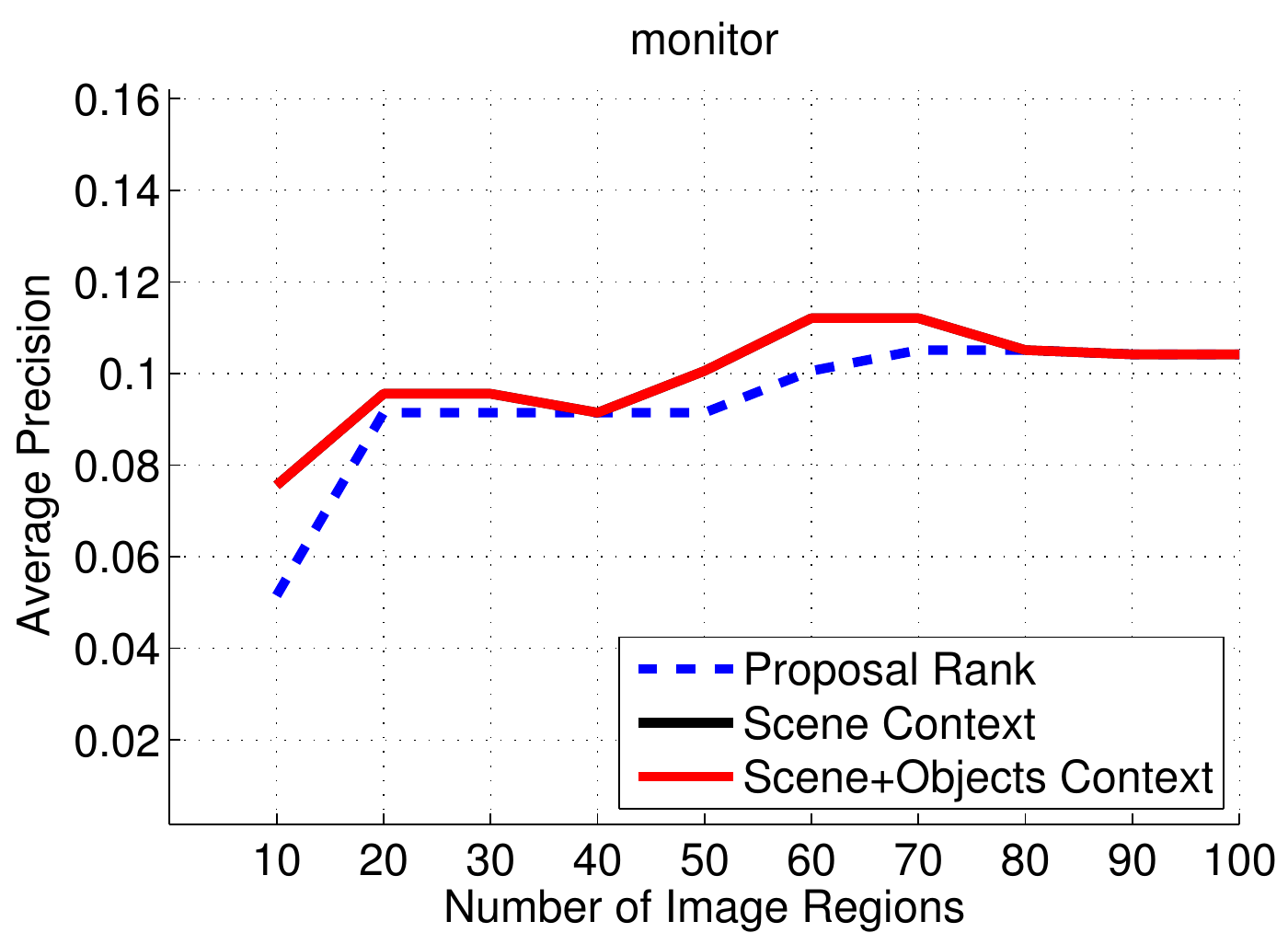} 
\includegraphics[width=1.2in]{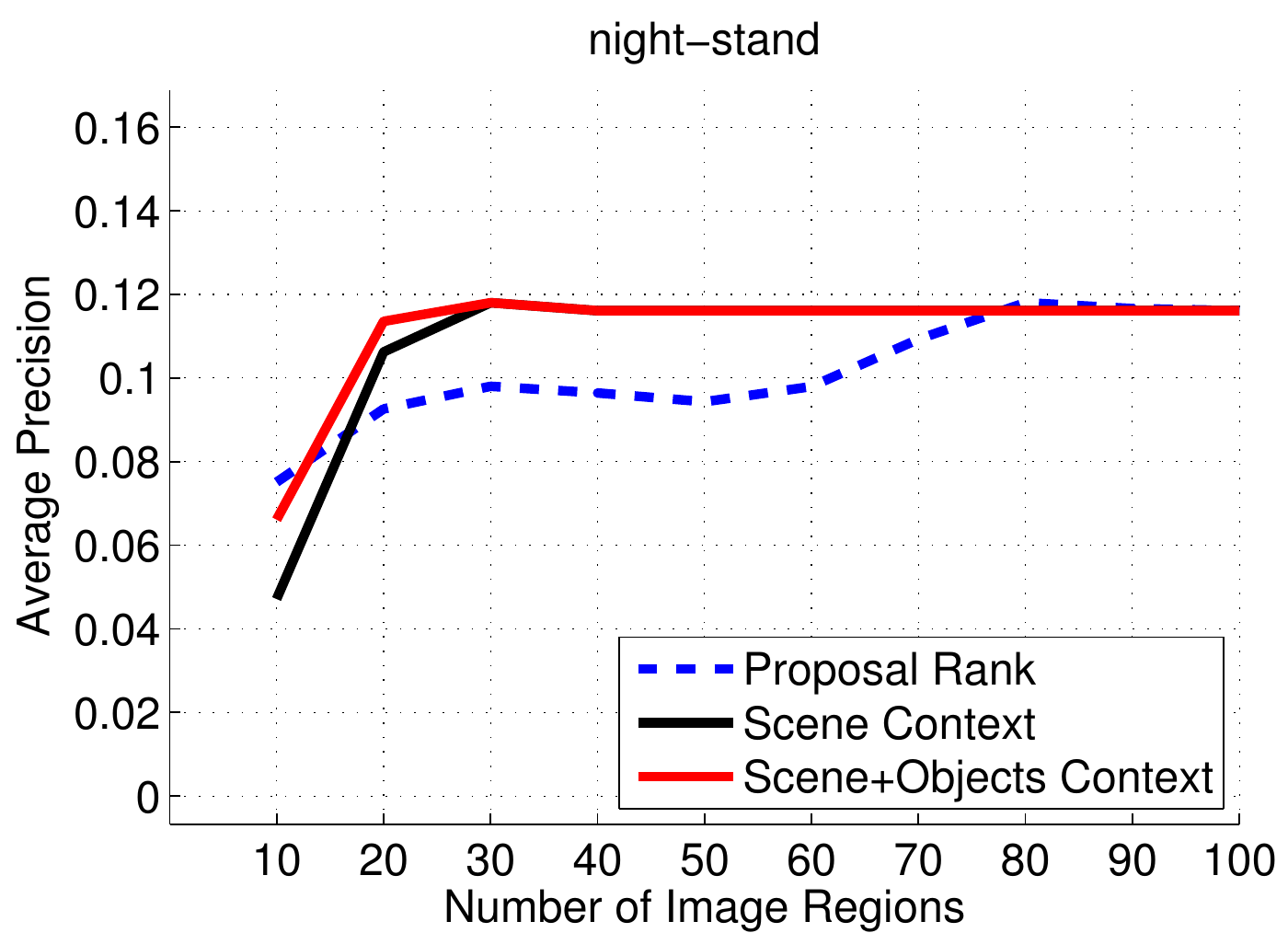} 
\includegraphics[width=1.2in]{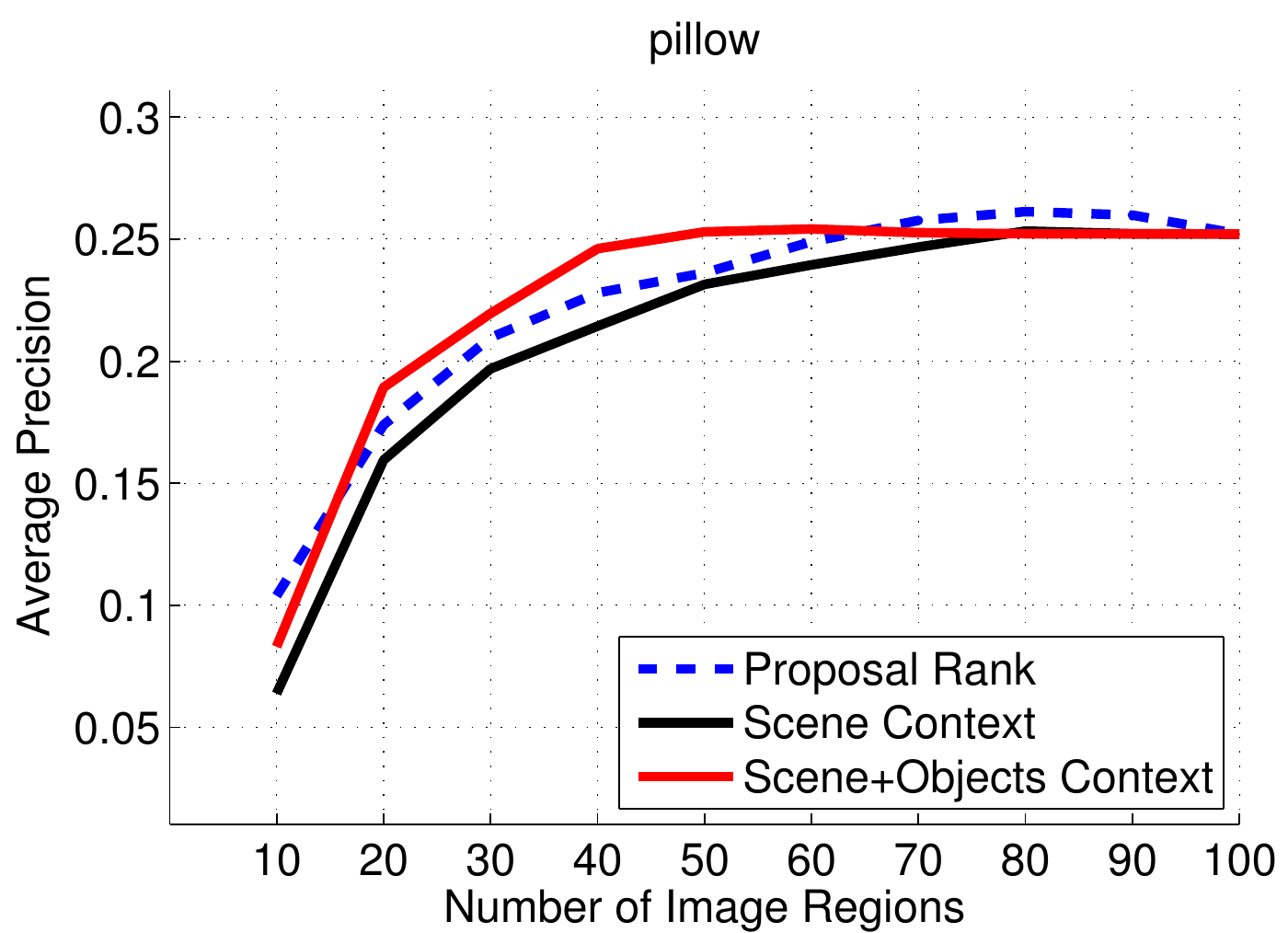} 
\includegraphics[width=1.2in]{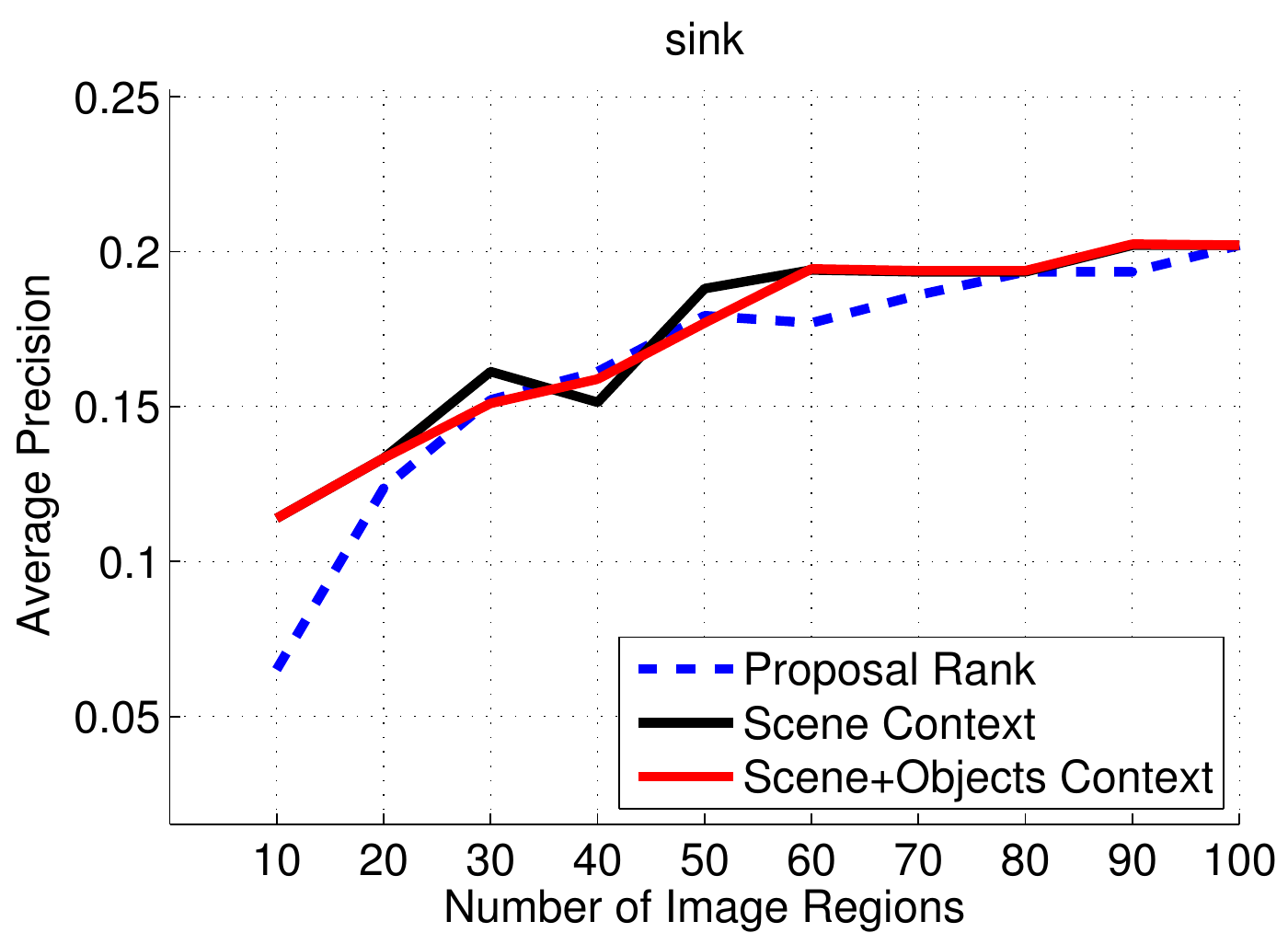} 
\includegraphics[width=1.2in]{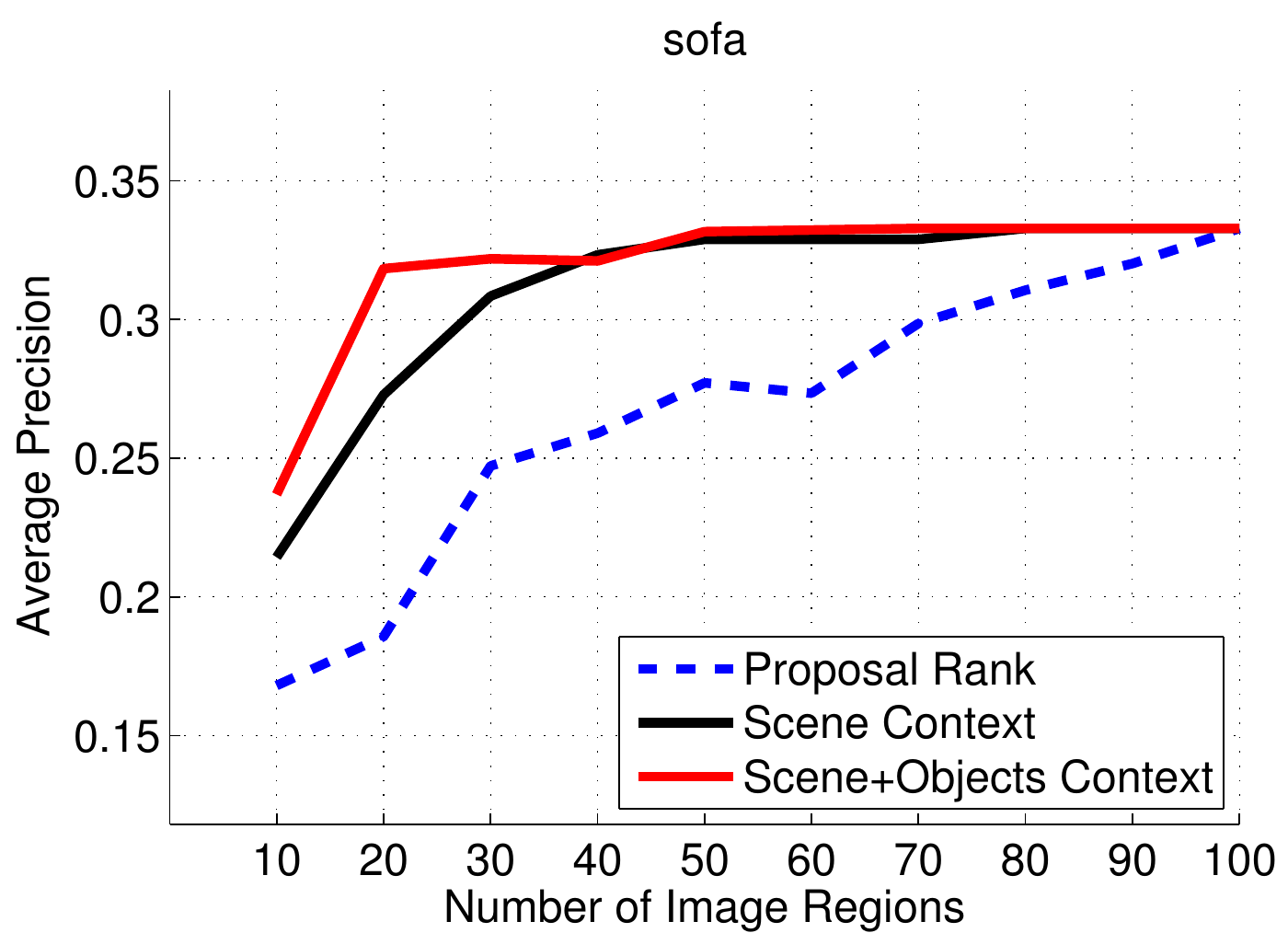} 
\includegraphics[width=1.2in]{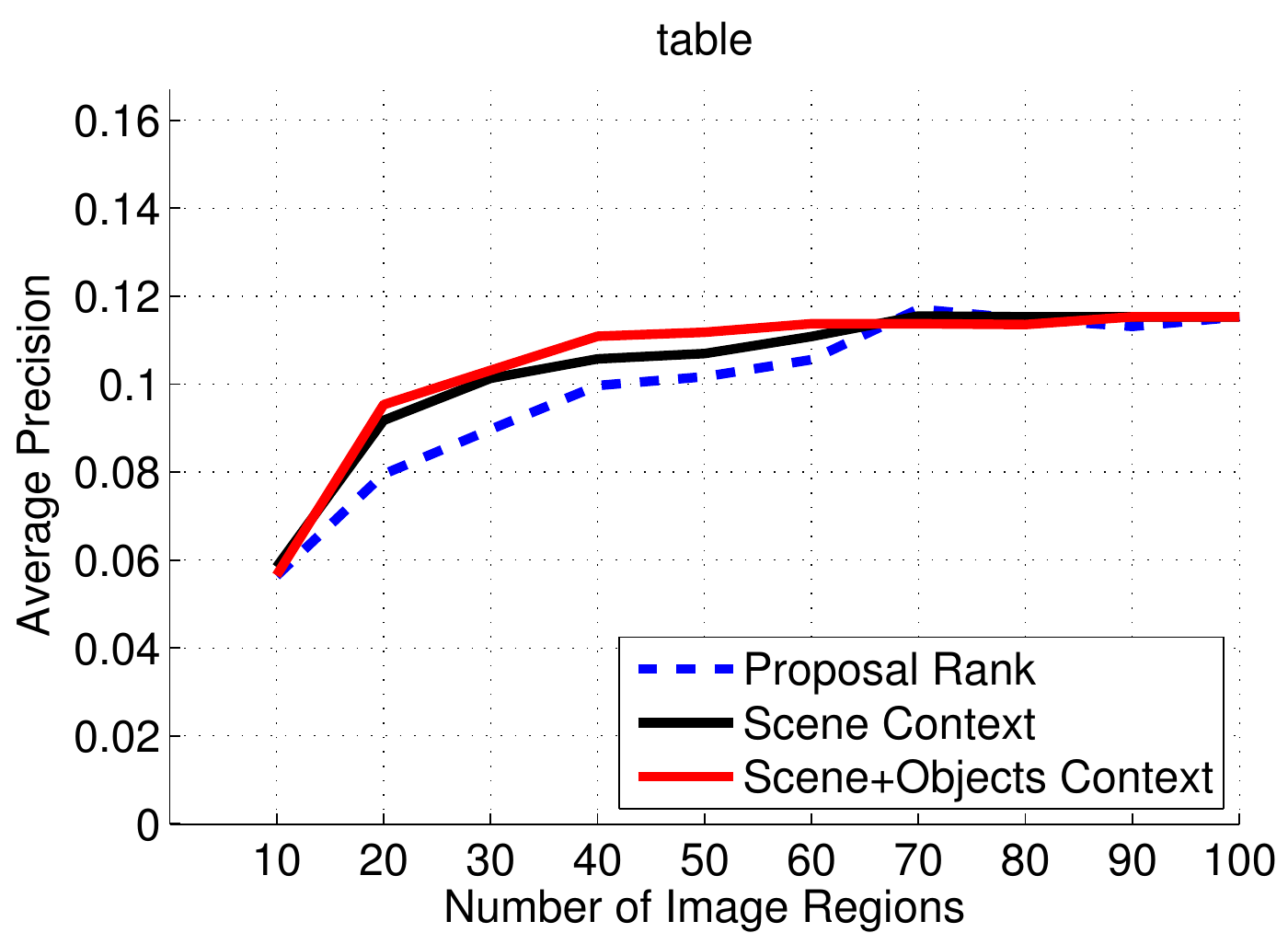} 
\includegraphics[width=1.2in]{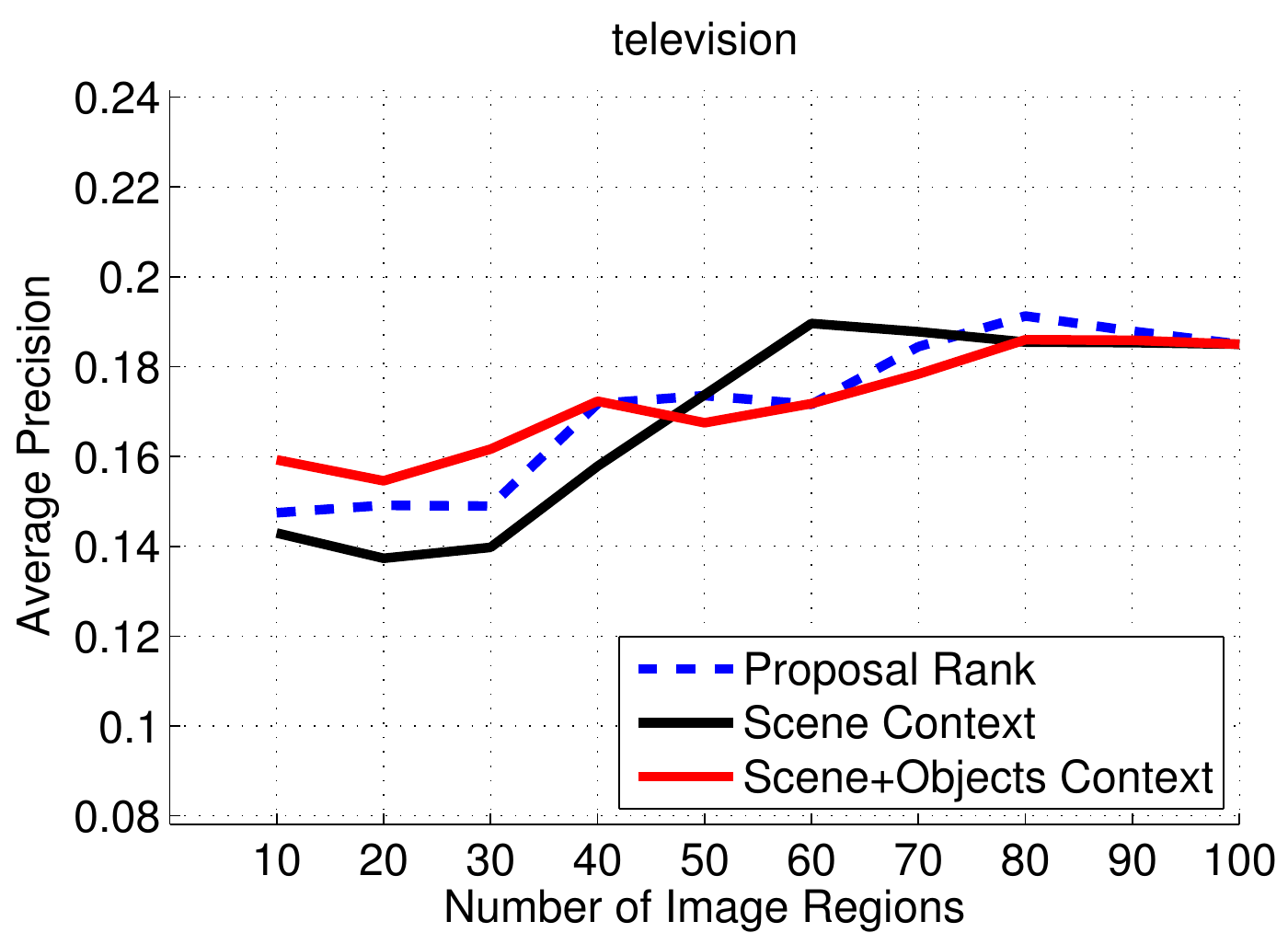} 
\includegraphics[width=1.2in]{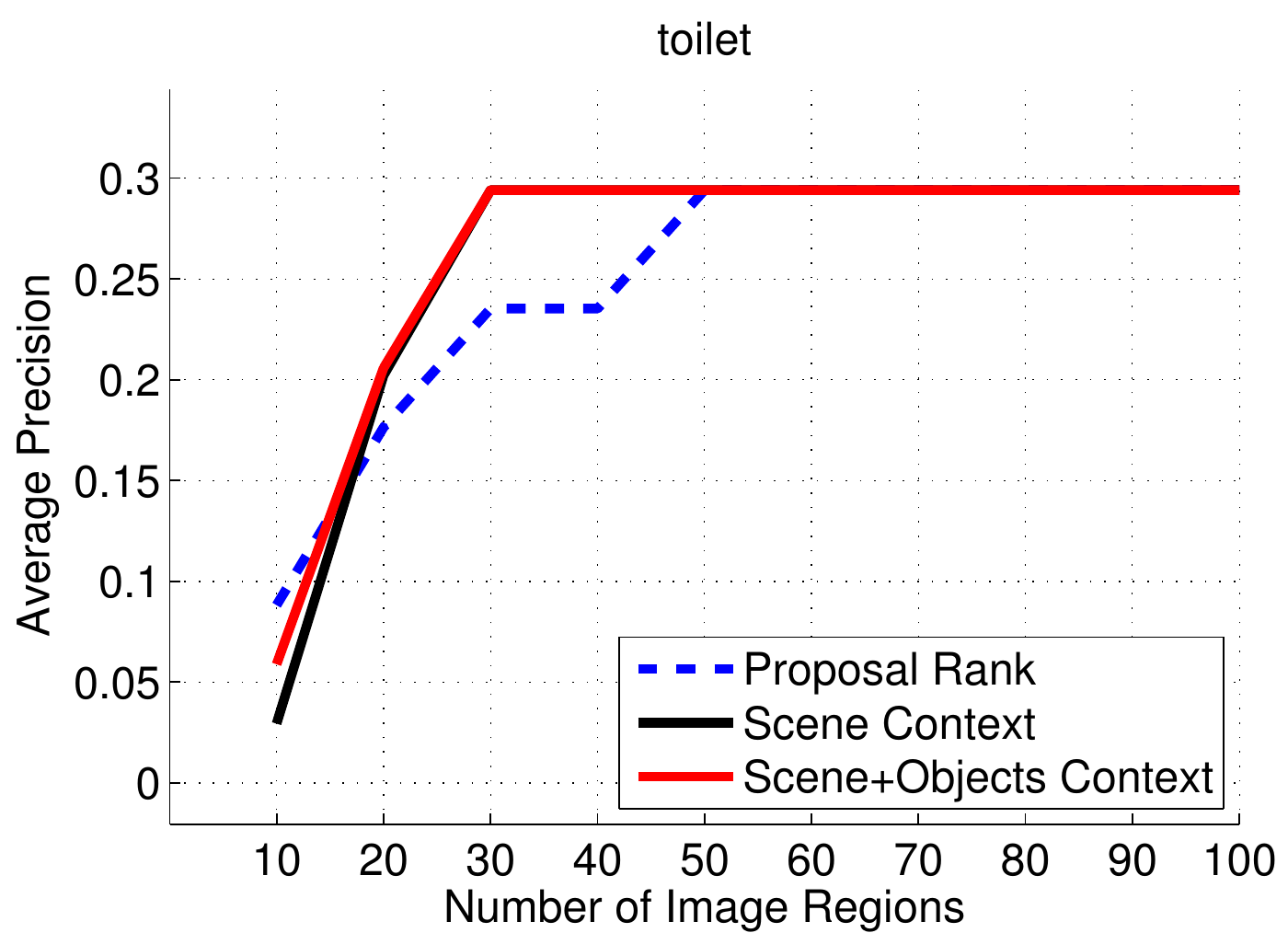} 
\caption{\textbf{Average Precision (AP) vs. number of processed regions.} A classifier trained for a query class with unary scene context features alone can achieve a significantly high average precision by processing very few regions. Classes like \textit{bed, nightstand} and \textit{sofa} need only 20-25\% of the regions when compared to the proposal ranking sequence. A search strategy trained for a query class using both object-object context and scene-context features further improves the performance for classes like \textit{counter, lamp, pillow} and \textit{sofa}. While the plots show sequential processing of all 100 regions, the stopping criterion for practical situations can be chosen based on the number of regions at which we obtain the maximum AP.}
\vspace{-0.2in}
\label{Fig:f1score}
\end{figure*}

Given a sequence of processed image regions, we measure the performance by the average precision (AP) of object detection performance versus the number of regions processed. Specifically, we measure the average precision at intervals of 10 image regions until we reach 100 image regions. Since our goal is to search for an object of a query class, the sequence of regions produced by our sequential exploration technique is different for different query classes. Figure \ref{Fig:f1score} shows average precision as the number of processed regions increases. Each figure compares various sequences produced for a particular query class. The baseline technique we compare with is the rank sequence obtained from the region proposal technique. The sequence is usually rank diversified and not necessarily sorted by objectness scores. Since the region proposal technique is not aware of the query class, it produces only one sequence for an image. 

\begin{figure}[t]
\addtolength{\subfigbottomskip}{-0.1in}
\centering
\footnotesize
\begin{tabu} to \textwidth {X[c]X[c]X[c]X[c]}
Groundtruth & Proposal Rank & Scene Context & Scene+Objects Context
\end{tabu}
\subfigure[Searching for chair. Number of regions processed = 15]{
\includegraphics[width=0.24\linewidth, trim=0cm 1cm 0cm 1cm,clip=true]{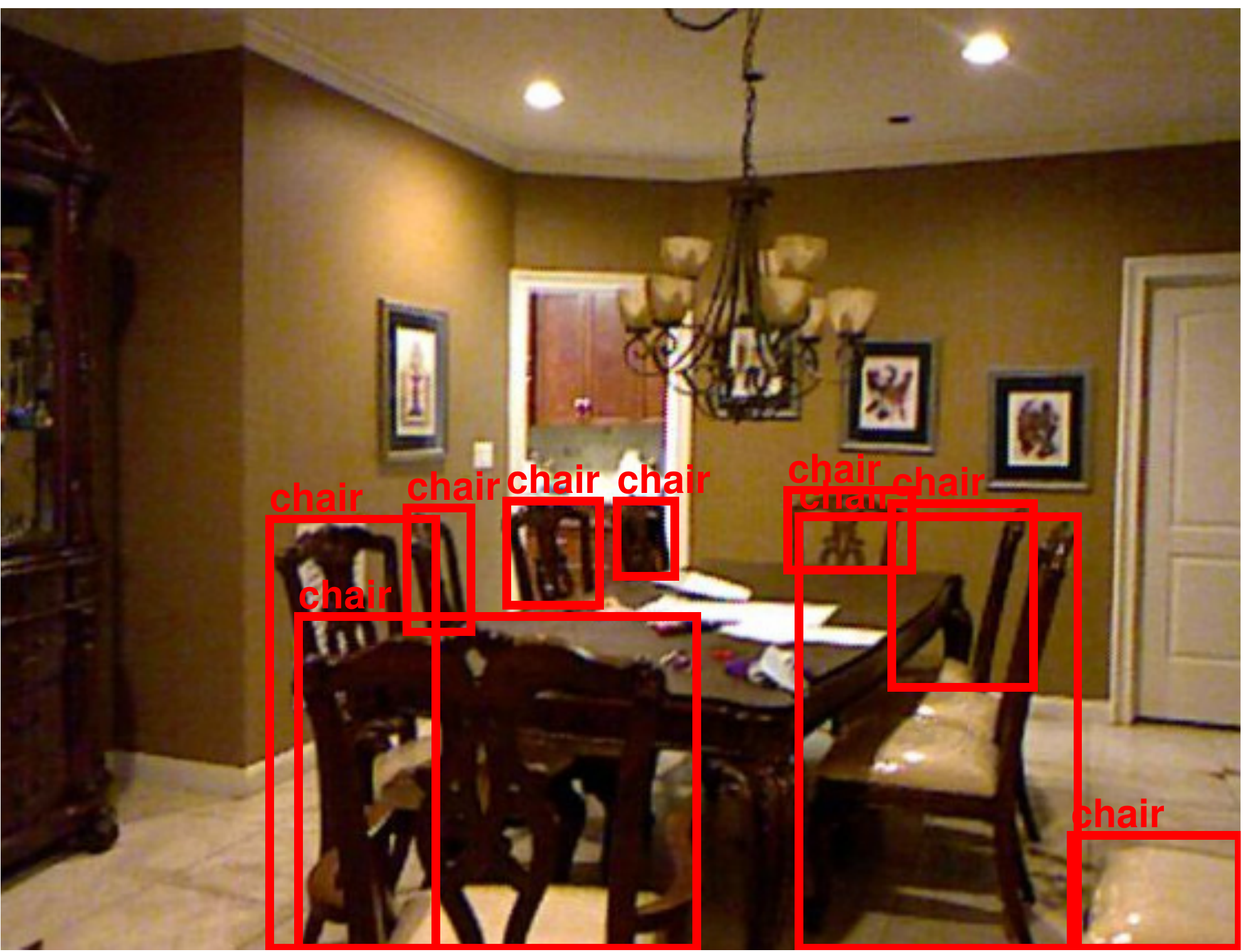}
\includegraphics[width=0.24\linewidth, trim=0cm 1cm 0cm 1cm,clip=true]{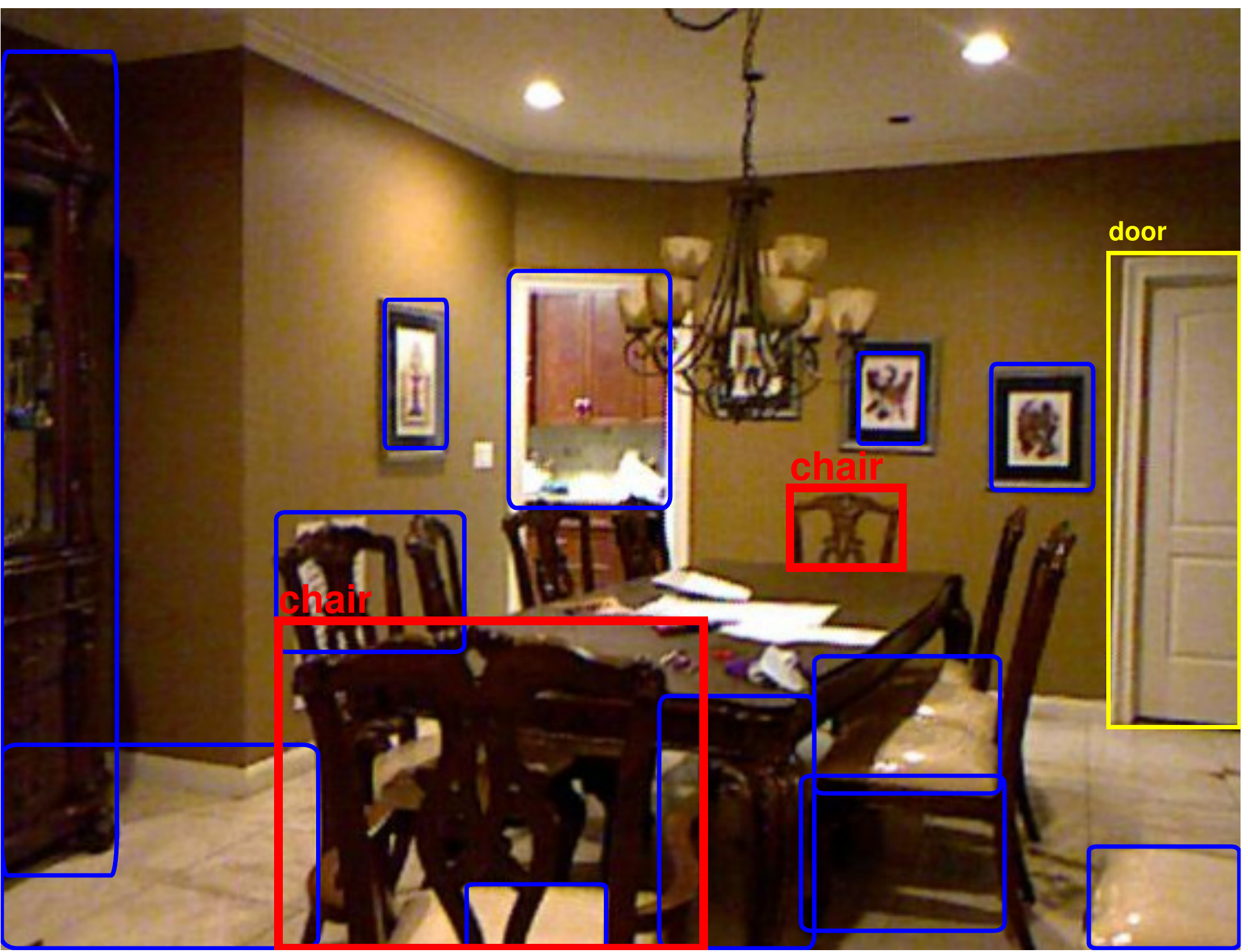}
\includegraphics[width=0.24\linewidth, trim=0cm 1cm 0cm 1cm,clip=true]{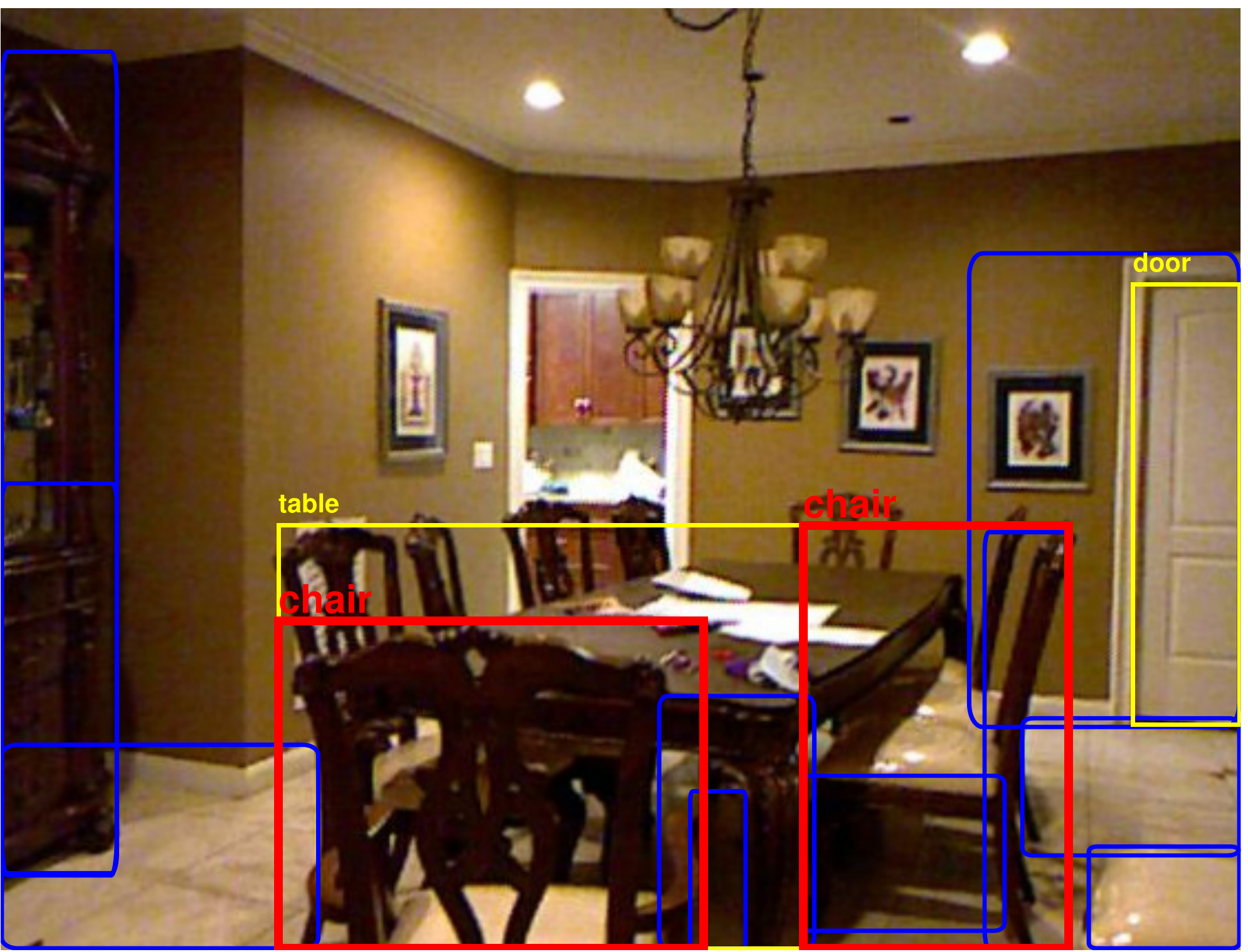}
\includegraphics[width=0.24\linewidth, trim=0cm 1cm 0cm 1cm,clip=true]{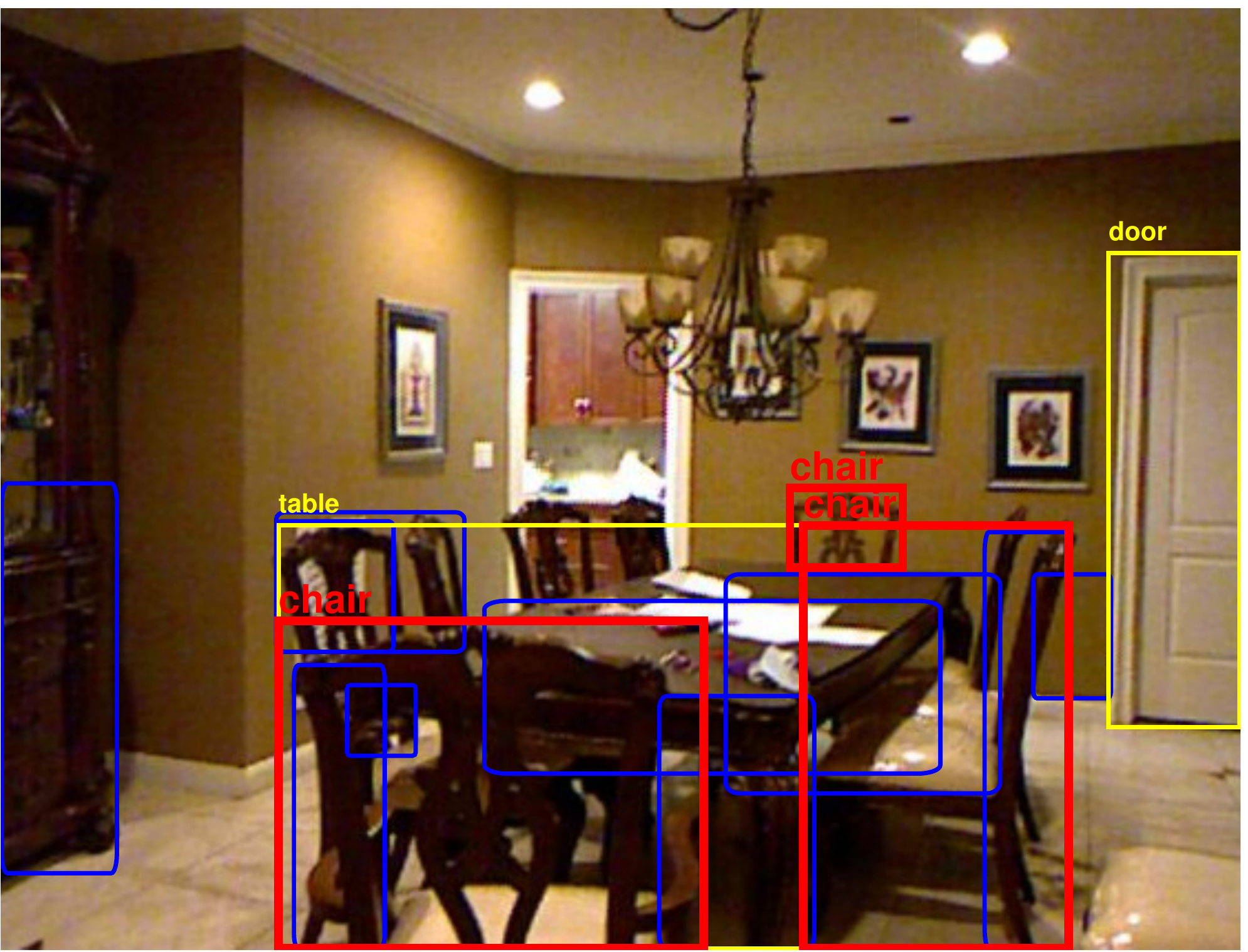}}

\subfigure[Searching for lamp. Number of regions processed = 35]{
\includegraphics[width=0.24\linewidth, trim=0cm 1cm 0cm 1cm,clip=true]{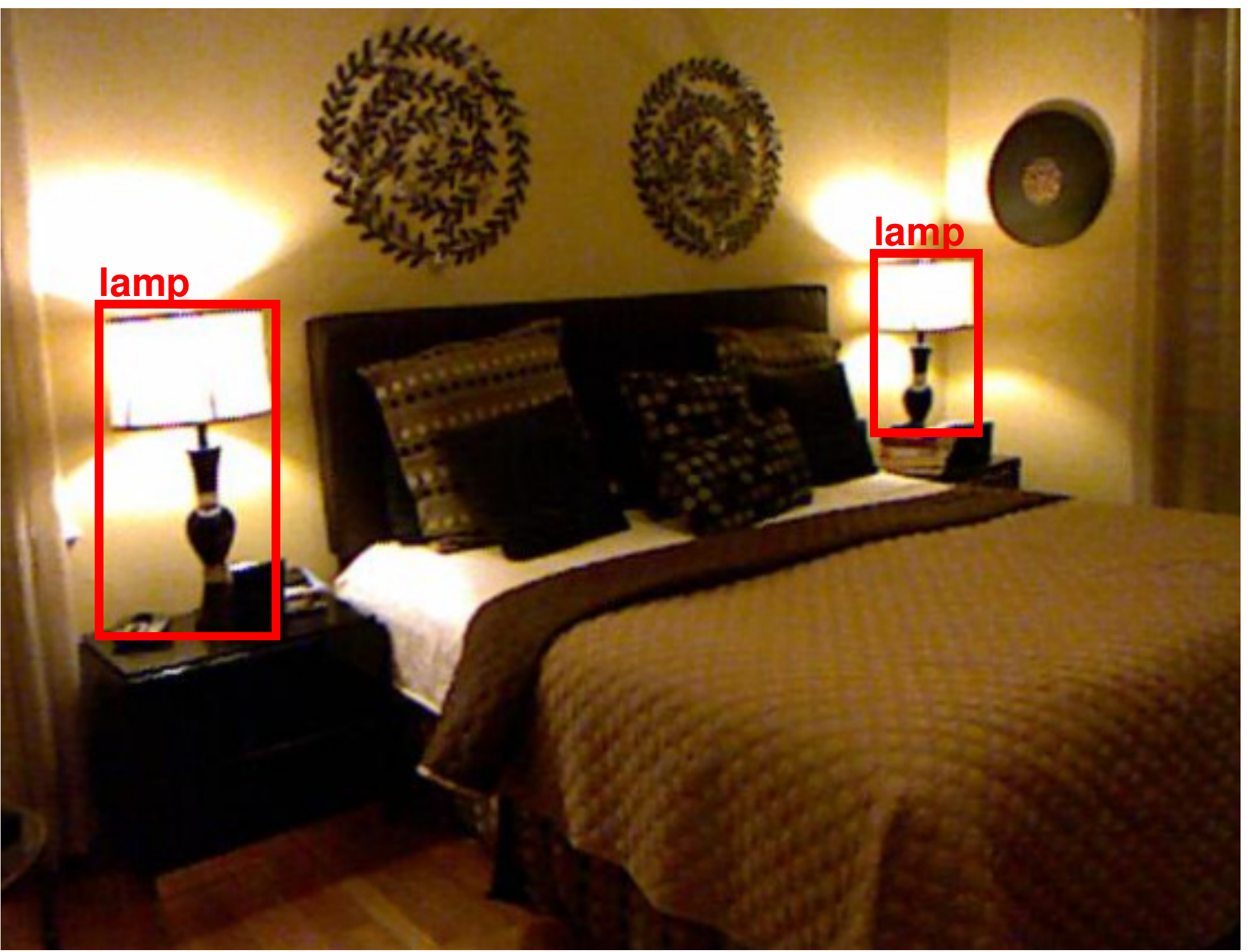}
\includegraphics[width=0.24\linewidth, trim=0cm 1cm 0cm 1cm,clip=true]{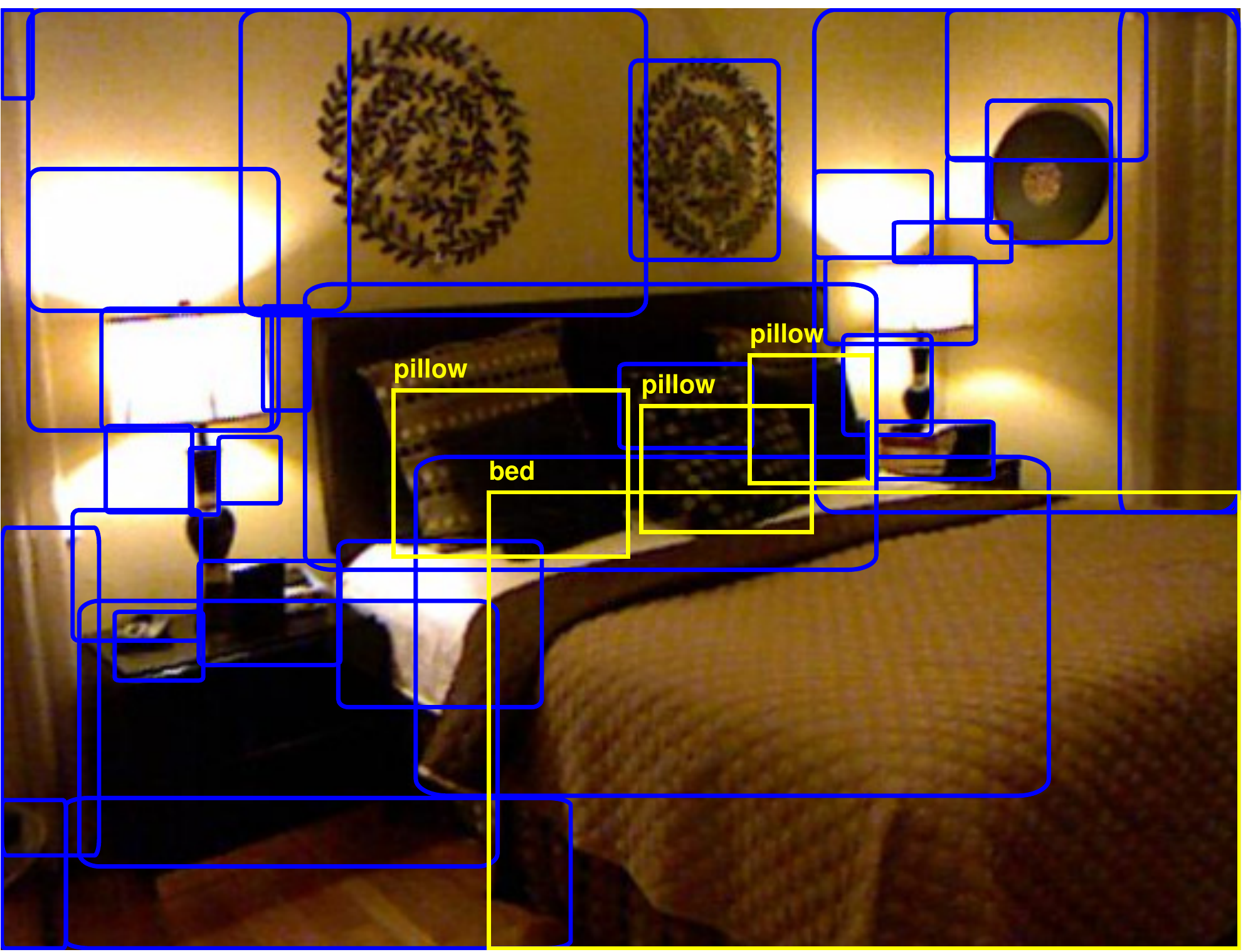}
\includegraphics[width=0.24\linewidth, trim=0cm 1cm 0cm 1cm,clip=true]{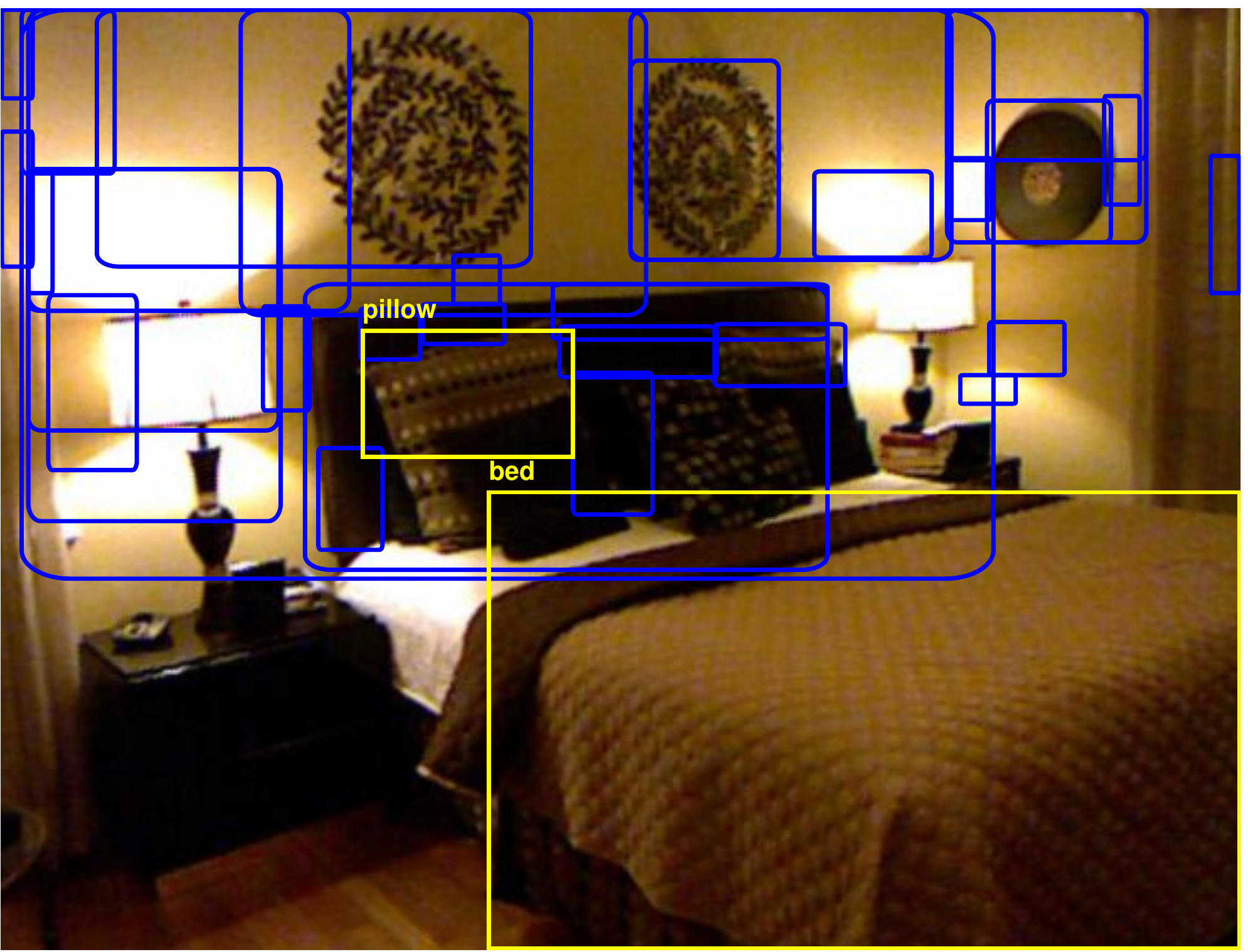}
\includegraphics[width=0.24\linewidth, trim=0cm 1cm 0cm 1cm,clip=true]{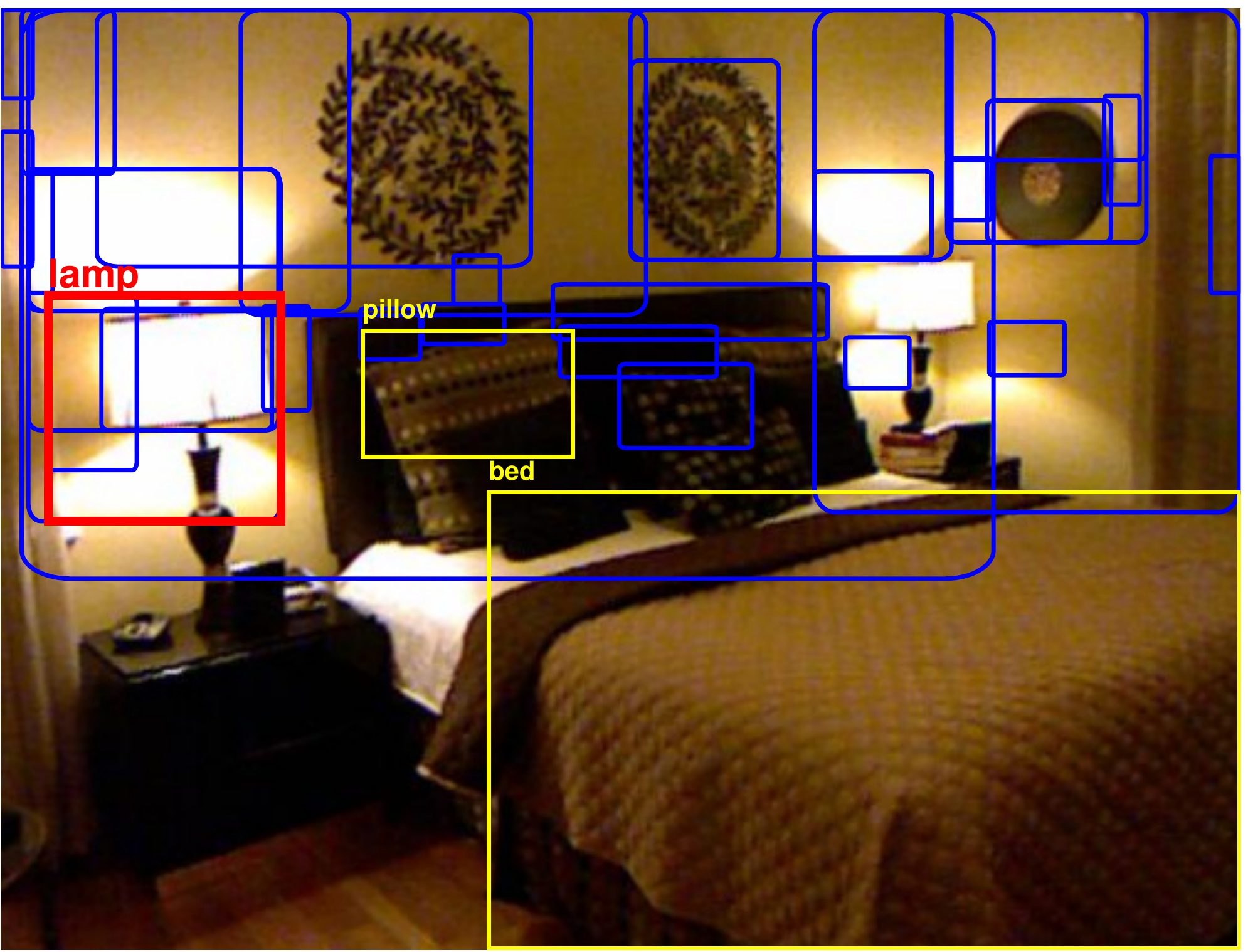}}

\subfigure[Searching for pillow. Number of regions processed = 15]{
\includegraphics[width=0.24\linewidth, trim=0cm 1cm 0cm 1cm,clip=true]{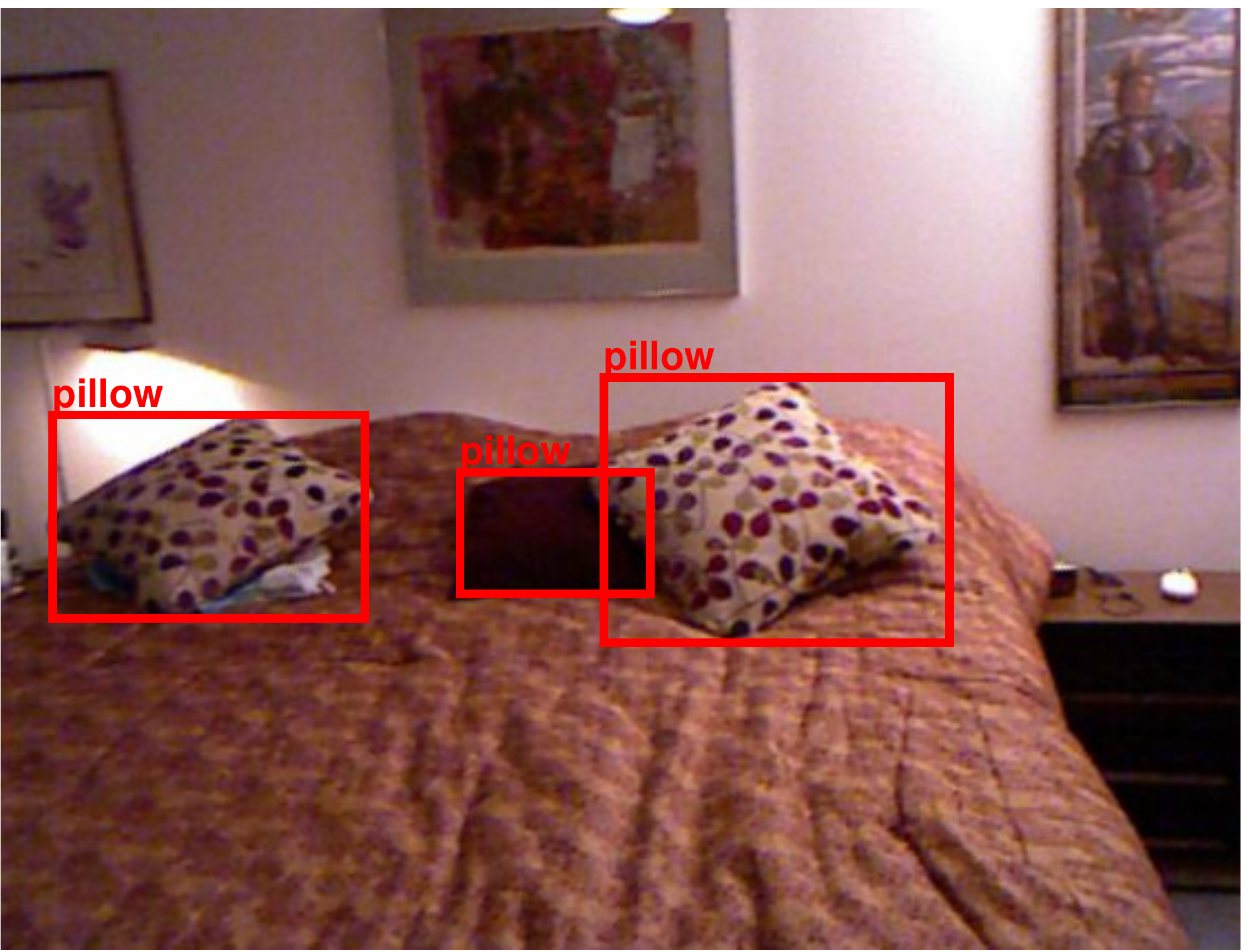}
\includegraphics[width=0.24\linewidth, trim=0cm 1cm 0cm 1cm,clip=true]{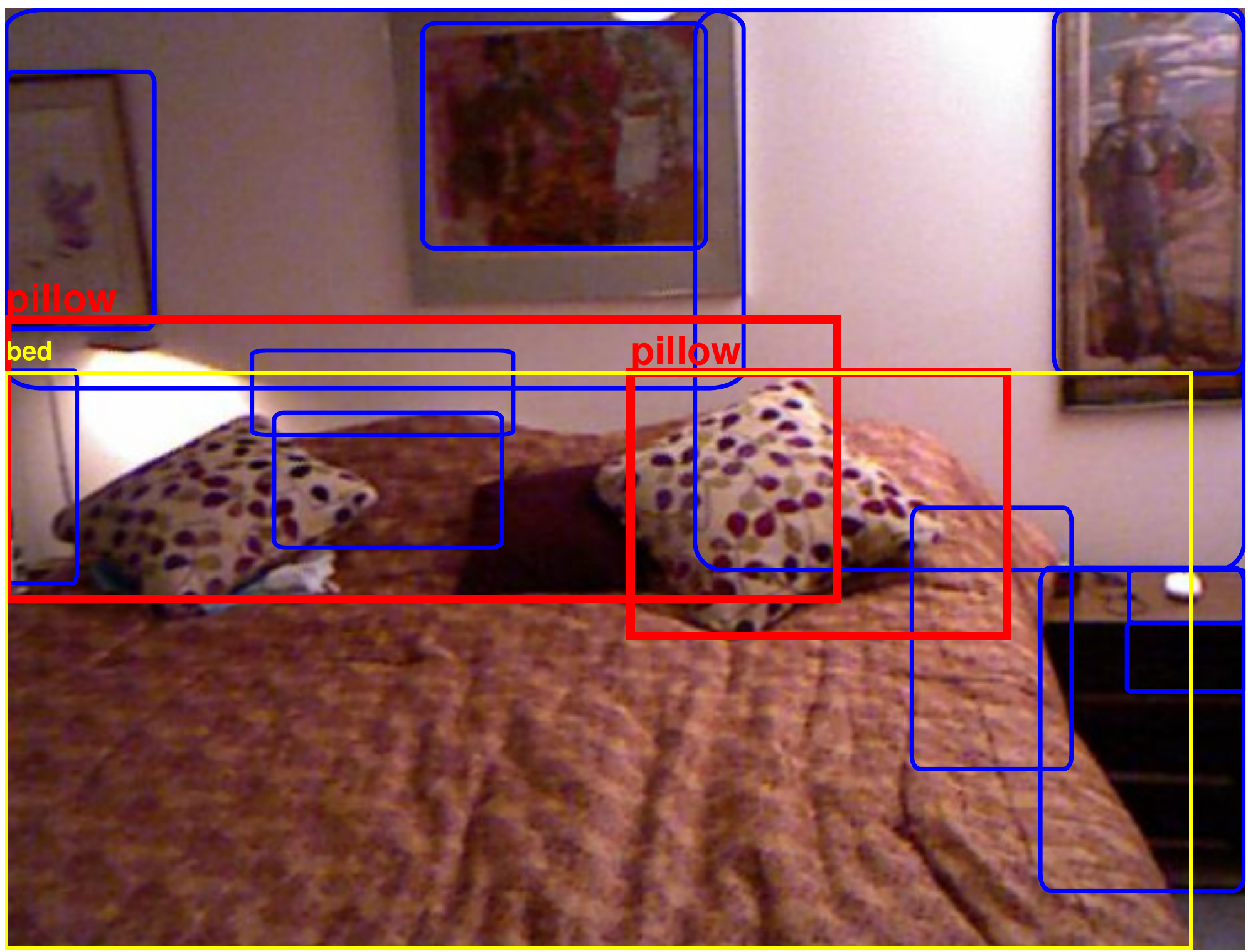}
\includegraphics[width=0.24\linewidth, trim=0cm 1cm 0cm 1cm,clip=true]{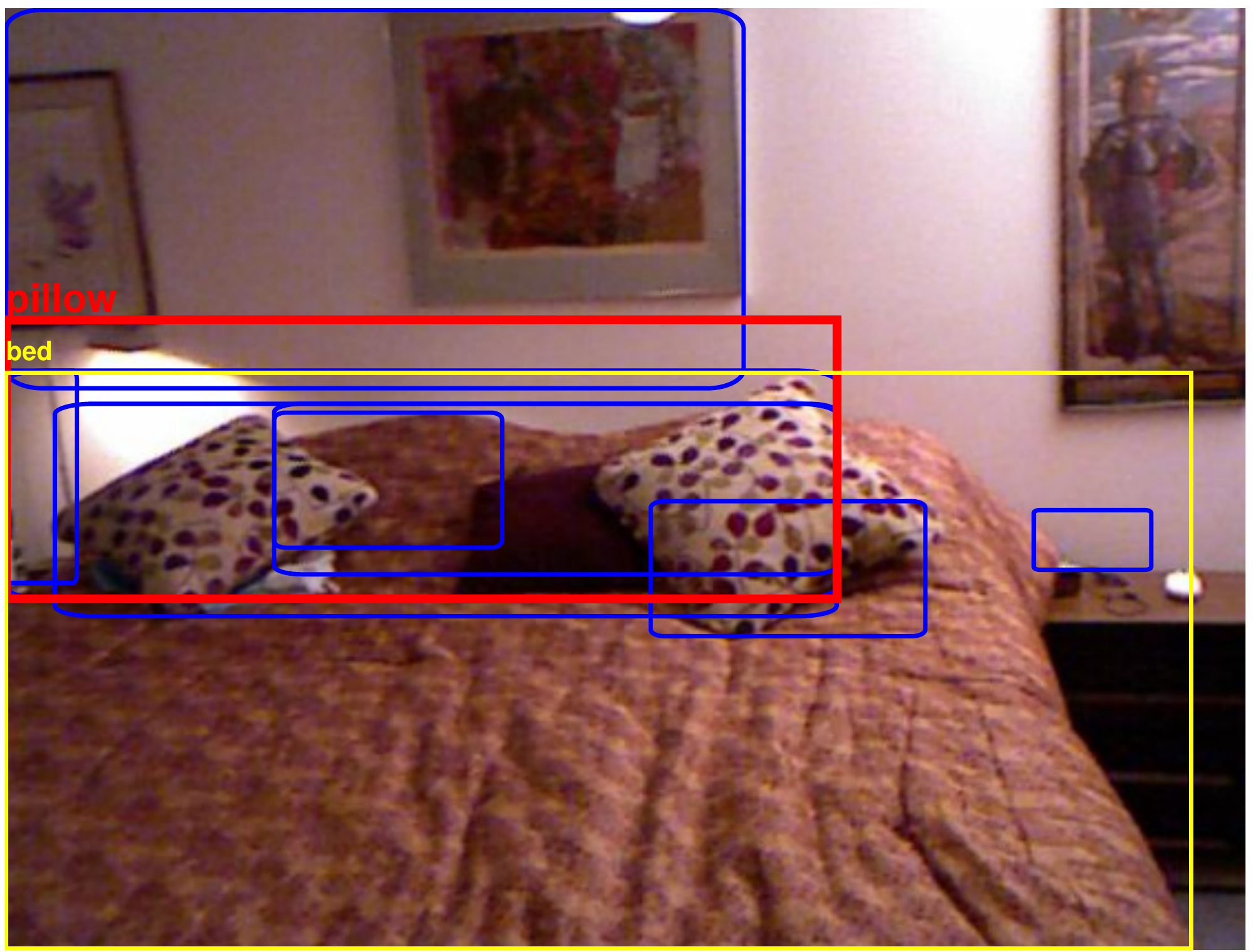}
\includegraphics[width=0.24\linewidth, trim=0cm 1cm 0cm 1cm,clip=true]{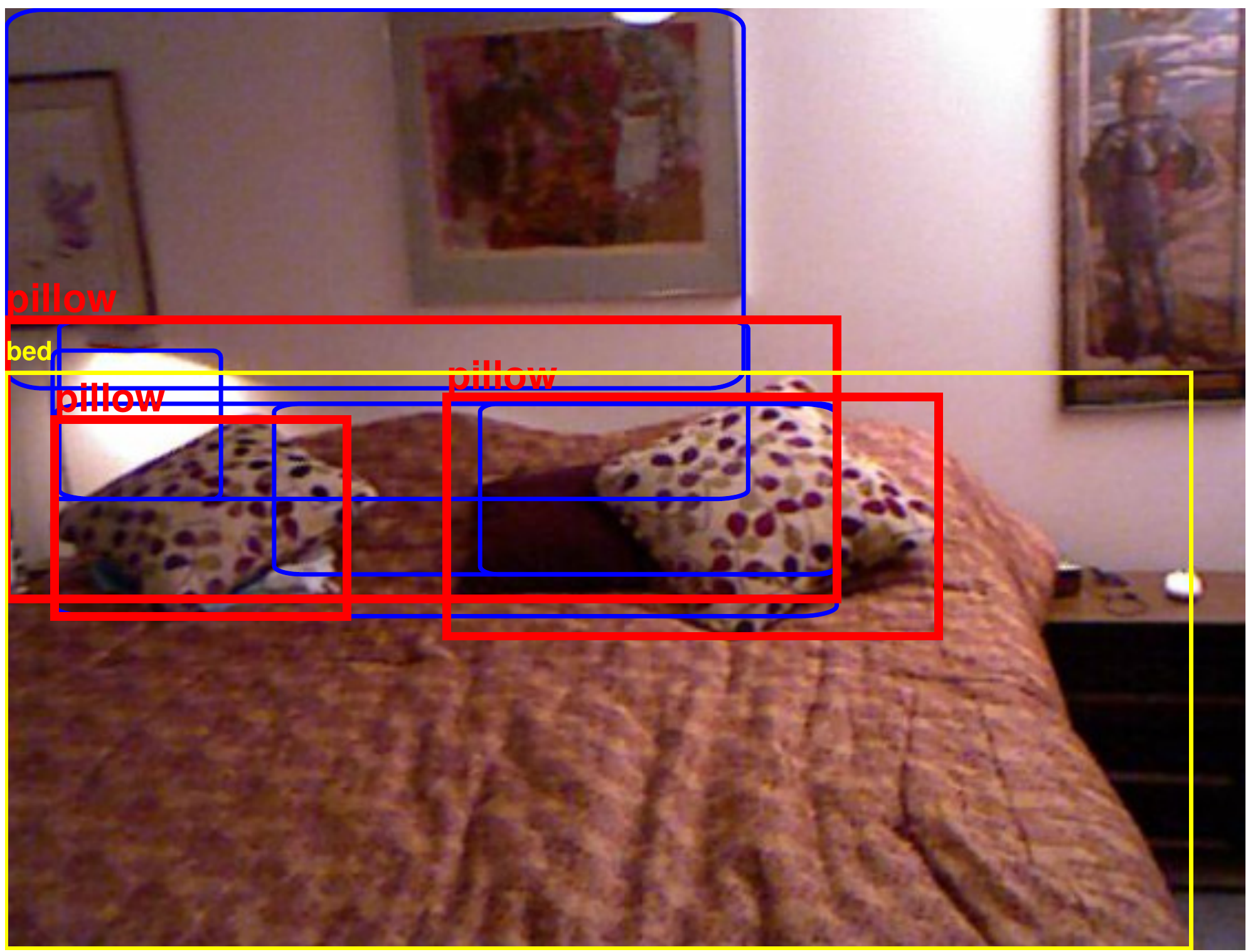}}

\subfigure[Searching for sofa. Number of regions processed = 20]{
\includegraphics[width=0.24\linewidth, trim=0cm 1cm 0cm 1cm,clip=true]{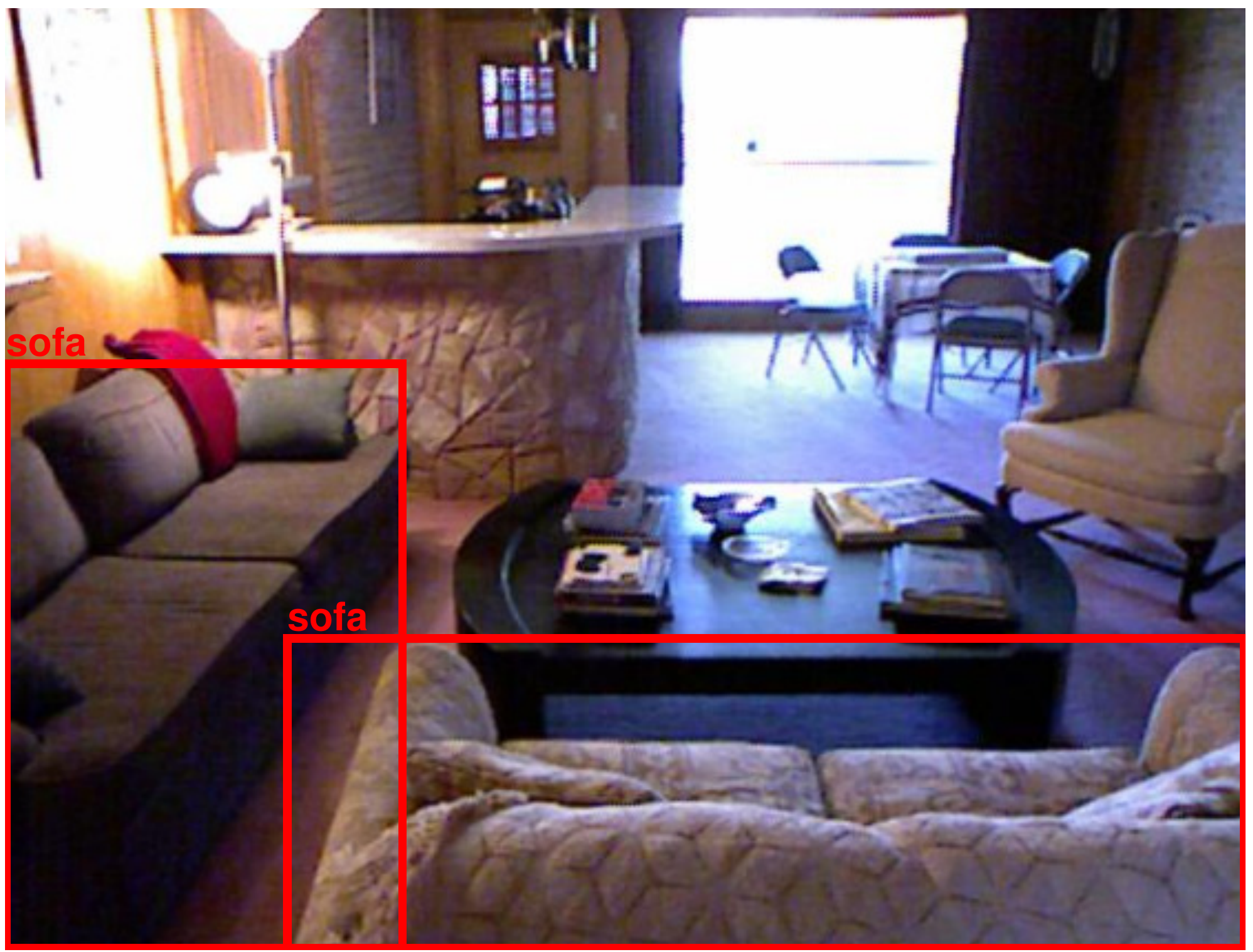}
\includegraphics[width=0.24\linewidth, trim=0cm 1cm 0cm 1cm,clip=true]{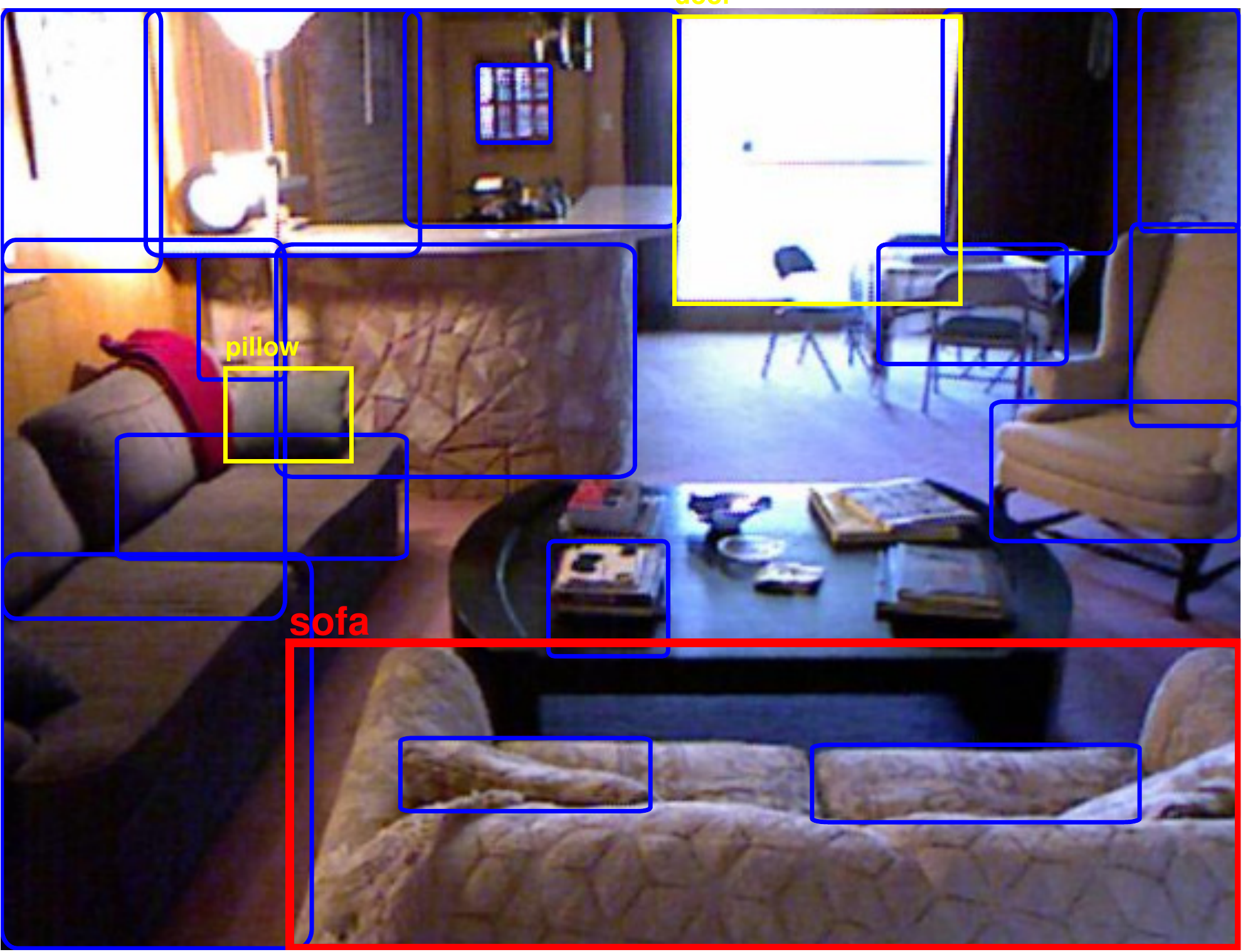}
\includegraphics[width=0.24\linewidth, trim=0cm 1cm 0cm 1cm,clip=true]{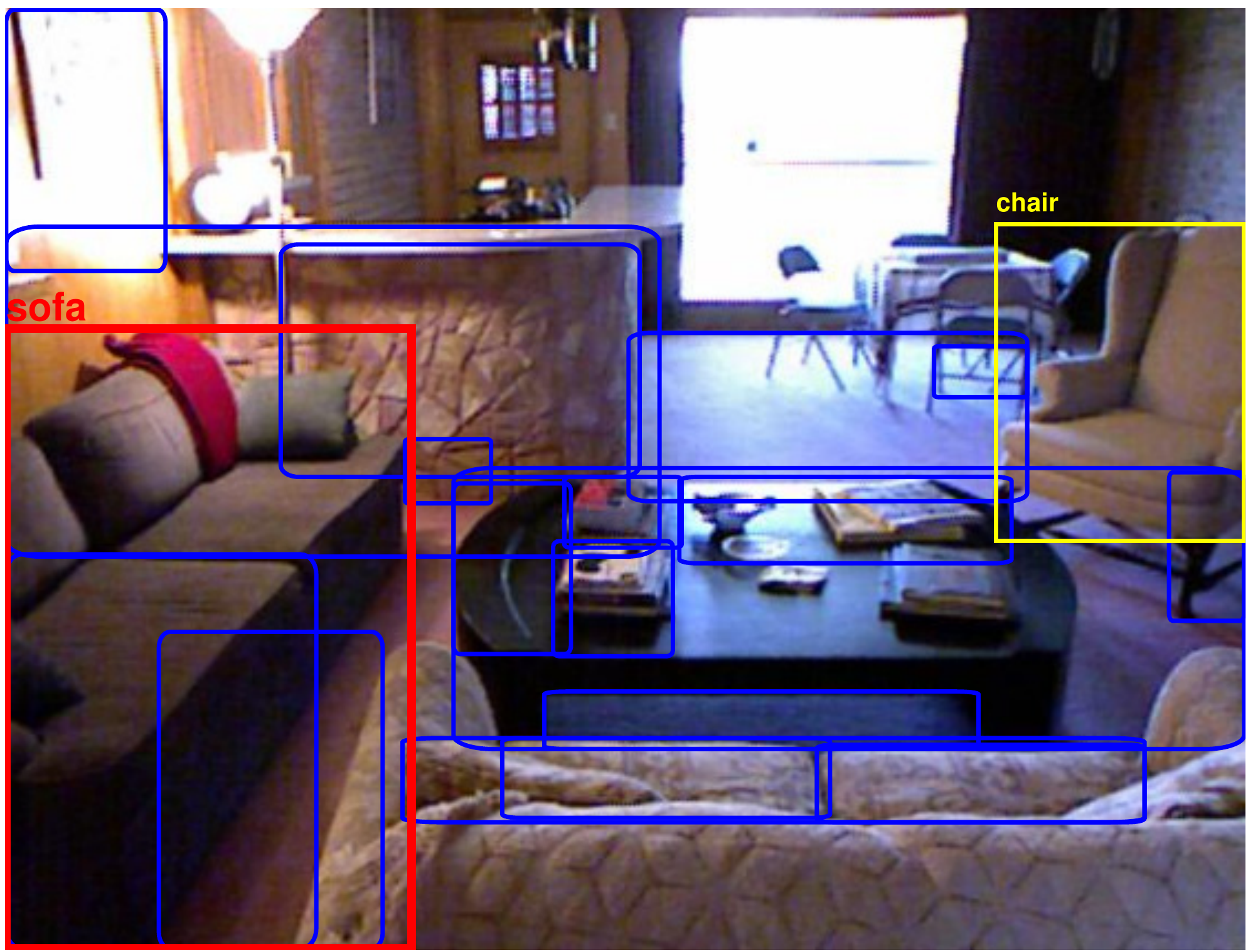}
\includegraphics[width=0.24\linewidth, trim=0cm 1cm 0cm 1cm,clip=true]{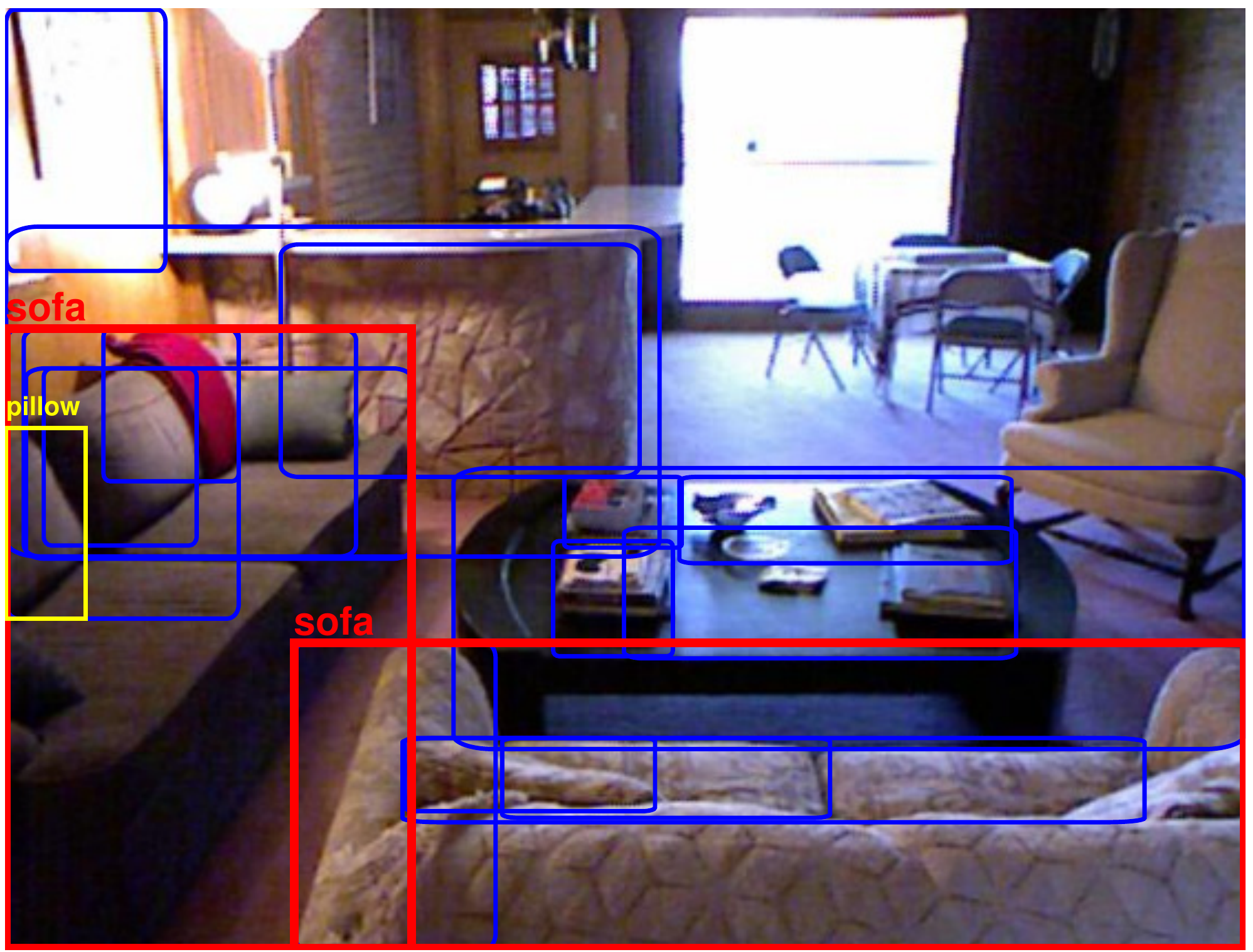}}

\caption{\textbf{Search results for different queries.} We compare three strategies - ranked sequence obtained from the region proposal technique (unaware of query class), ranked sequence obtained from a classifier trained for a query class using scene context features alone and sequence produced by a search strategy trained for a query class using both scene context and object-object context features. Red boxes indicate regions labeled as query class, yellow boxes indicate regions other than the query class and blue boxes indicate regions labeled as background. The images show a state in the search sequence of different methods at a certain number of regions processed. We can see that our strategy which uses both scene context and object-object context can locate an object of the query class earlier than the other methods.}
\vspace{-0.15in}
\label{Fig:SearchResults}
\end{figure}

First, we train a classifier with the query class as the label and just the unary scene context features (see Sec. \ref{Sec:SeqExplore}). Since the scene context features do not change based on the regions explored, we obtain scores from this classifier in a single step. The scores are then used to rank the regions to obtain a sequence. Our results indicate that the scene context features alone can achieve a significantly high average precision by using very few regions - for some classes (Ex: \textit{bed, nightstand, sofa}) almost 20-25\% of the regions when compared with the proposal ranking sequence. Next, we perform a sequential search using strategies trained with object-object context features along with the scene context features. The results indicate that for classes like \textit{counter, lamp, pillow} and \textit{sofa}, object-object context improves the average precision over using just the scene context features. While we see improvement in the dresser class as well, the number of test samples are too few to determine the significance of the improvement. Figure \ref{Fig:SearchResults} shows examples of search results for different query classes. The examples show that our strategy which uses both scene context and object-object context can locate objects of the query class earlier than the other methods. 

The supplementary material contains plots that compares our technique trained with a randomly selected background subset against our technique trained with the determinant maximization based background subset. The plots show that our determinant maximization based subset selection technique performs better or equally well with the random subset on most of the classes. But the main advantage of our subset selection technique is the repeatability of experiments unlike the one with random subset selection.

\textbf{Computation time:} On a single core of an Intel 4.0GHz processor, it takes only 20ms on average for the search process in an image with 100 region proposals. The time taken for extracting CNN features is 5ms per region on a GPU. Since our results show that for most of the classes we can achieve a high average precision at around 25 to 50 regions instead of evaluating all 100 regions, the total time taken for feature extraction and search overhead is 0.145s and 0.27s for 25 and 50 regions respectively. This shows that the search overhead is negligible compared to the total time and the reduction in number of regions directly translates to 2 to 4 times speedup in computation time while still achieving a high average precision. The time for context feature extraction is negligible because the necessary information is already extracted by the region proposal module.

\section{Conclusion}
We have proposed a search technique for detecting objects of a particular class in an image by processing as few image regions as possible. The search strategy is framed as a Markov decision process learned using an imitation learning algorithm, which sequentially explores regions based on structure in the scene. Our experiments show that unary scene context features of regions can alone achieve a significantly high average precision after processing only 20-25\% of the regions for classes like \textit{bed, night-stand} and \textit{sofa}. By incorporating object-object context, the performance is further improved for classes like \textit{counter, lamp, pillow} and \textit{sofa}. Our sequential search process adds a negligible overhead when compared to the time spent on extracting CNN features, hence the reduction in number of regions leads directly to a gain in computation speed of the object detection process.

\section{Acknowledgements}
This research was supported by contract N00014-13-C-0164 from the Office of Naval Research through a subcontract from the United Technologies Research Center. The GPUs used in this research were generously donated by the NVIDIA Corporation. We thank Hal Daum\'{e} III for helping us with the Vowpal Wabbit code and Saurabh Gupta for helping us with his RCNN-depth code.

\bibliography{../../../BibTeX_Colln/Object_Spotting}
\end{document}